%% file: MAIN.tex
\documentclass[format=manuscript,screen=true,review=false,nonacm=true]{acmart} %
\AtBeginDocument{%
  }

\PassOptionsToPackage{dvipsnames}{xcolor} %
\usepackage{subcaption} %
\usepackage[inline]{enumitem} %

\input{macros}

\begin{document}

\makeatletter
\fancypagestyle{standardpagestyle}{%
  \fancyhf{}%
  \renewcommand{\headrulewidth}{\z@}%
  \renewcommand{\footrulewidth}{\z@}%
  \fancyhead[LE]{\if@ACM@printfolios\thepage\fi}%
  \fancyhead[RO]{\if@ACM@printfolios\thepage\fi}%
  \fancyhead[RE]{\@shortauthors}%
  \fancyhead[LO]{\shorttitle}%
  \fancyfoot[RO,LE]{\footnotesize Preprint}%
}
\fancypagestyle{firstpagestyle}{%
  \fancyhf{}%
  \renewcommand{\headrulewidth}{\z@}%
  \renewcommand{\footrulewidth}{\z@}%
  \fancyfoot[RE,LO]{\footnotesize Preprint}%
  \fancyfoot[RO,LE]{\if@ACM@printfolios\small\thepage\fi}%
}
\pagestyle{standardpagestyle}
\makeatother

\title{Plasticity Loss in Deep Reinforcement Learning: A Survey}

\author{Timo Klein}
\email{timo.klein@univie.ac.at}
\orcid{0000-0002-0195-7897}
\affiliation{%
  \institution{Faculty of Computer Science, University of Vienna}
  \city{Vienna}
  \country{Austria}
}
\affiliation{%
  \institution{Doctoral School Computer Science, University of Vienna}
  \city{Vienna}
  \country{Austria}
}

\author{Christoph Luther}
\email{cph.luther@gmail.com}
\affiliation{%
  \institution{Independent Researcher}
  \city{Vienna}
  \country{Austria}
}

\author{Manus McAuliffe}
\authornote{Work done prior to joining Microsoft AI.}
\email{manusmcauliffe123@gmail.com}
\affiliation{%
  \institution{Microsoft AI}
  \city{London}
  \country{United Kingdom}
}

\author{Lukas Miklautz}
\orcid{0000-0002-2585-5895}
\affiliation{%
  \institution{Department of Machine Learning and Systems Biology, Max Planck Institute
of Biochemistry}
  \city{Martinsried}
  \country{Germany}
}

\author{Claudia Plant}
\orcid{0000-0001-5274-8123}
\author{Sebastian Tschiatschek}
\orcid{0000-0002-2592-0108}
\affiliation{%
  \institution{Faculty of Computer Science, University of Vienna}
  \city{Vienna}
  \country{Austria}
}
\affiliation{%
  \institution{ds:UniVie}
  \city{Vienna}
  \country{Austria}
}

\renewcommand{\shortauthors}{Klein, Luther, McAuliffe, Miklautz, Plant, and Tschiatschek}

\input{sections/0abstract}



\maketitle

\input{sections/1Introduction}

\input{sections/2preliminaries}

\input{sections/3related_work}

\input{sections/4causes}

\input{sections/5mitigation}

\input{sections/6discussion_directions}

\input{sections/7conclusion}

\begin{acks}
This work has been funded in parts by the Vienna Science and Technology Fund (WWTF) [10.47379/ICT20058]. We thank Mateusz Ostaszewski for the insightful discussions that substantially improved our work. 
\end{acks}

\bibliographystyle{ACM-Reference-Format}
\bibliography{bib/plasticity,bib/rl,bib/other_ml,bib/random_intro_stuff,bib/related_work}

\end{document}

%% file: sections/0abstract.tex
\begin{abstract}%
Plasticity refers to a network's ability to adapt to changing data distributions, which is crucial for the successful training of deep reinforcement learning agents. Loss of plasticity causes performance plateaus and contributes to scaling failures, overestimation bias, and insufficient exploration. To deepen the understanding of plasticity loss, we propose a unified definition, examine its drivers and pathologies, and organize over 50 mitigation strategies into the first comprehensive taxonomy of the field. Our analysis shows gaps in current evaluation practices and reveals that general regularization techniques often outperform domain-specific interventions. Future research should prioritize understanding the mechanisms underlying plasticity loss.
\end{abstract}

%% file: sections/1Introduction.tex
\section{Introduction}\label{sec:introduction}

Deep Reinforcement Learning (RL) has recently seen many successes and breakthroughs: It beat the best human players in Go~\citep{AlphaGO} and Dota~\citep{OpenAIDota}, discovered new matrix multiplication algorithms~\citep{AlphaZeroMatrixMultiplication}, endowed language models with the ability to generate human-like replies for breaking the Turing test~\citep{ChatGPTTuringTest}, and allowed for substantial progress in robotic control~\citep{RubiksCubeHand}. Its capabilities to react to environmental changes and make near-optimal decisions in challenging sequential decision-making problems are likely crucial for any generally capable agent. Also, RL's capability to learn purely from trial-and-error mimics human learning, making it a natural paradigm for modeling learning in artificial agents~\citep{SuttonBartoRLIntroduction}.

Despite all the aforementioned successes, deep RL is still in its infancy, and current methods are often not yet reliable or mature. To reach high levels of performance, deep RL typically needs substantial tweaking and elaborate stabilization techniques that are notoriously difficult to get right: From replay buffers and target networks that stabilize temporal-difference learning~\citep{DQNOriginal}, to noise decorrelation and pessimistic value functions that address overestimation bias~\citep{DoubleQLearning,TD3OriginalPaper}, and finally to idiosyncratic optimizer settings and bespoke hyperparameter schedules that manage gradient pathologies~\citep{UnderstandingPlasticity, ResettingAdamMoments, BiggerBetterFaster}.

There are many reasons why this is the case: First and foremost, deep RL is inherently non-stationary, making it a substantially harder learning problem than supervised learning. Additionally, it suffers from its own optimization issues, such as under-exploration, sample correlation, and overestimation bias. Much recent work has been devoted to tackling these problems with increasingly elaborate algorithms, many of which aim to transfer insights from tabular RL to the deep RL setting~\citep{GeneralizedPessimismLearning, PseudocountExplorationDensityModels}.

\textbf{Recent work suggests that many of these issues --- overestimation bias~\citep{OverestimationOverfittingPlasticity}, scaling failures~\citep{simBa}, training instability~\citep{PQNParallelisedQNetwork} --- share a common root cause: the optimization difficulties that arise when training neural networks on non-stationary data.} This perspective has gained traction under the umbrella term \emph{plasticity loss}. In deep learning, plasticity refers to a network's ability to quickly adapt to new targets; plasticity loss characterizes a network state in which this ability has degraded. Because deep RL agents continuously update their policies and value estimates, they induce distribution shifts in their own training data. These are precisely the conditions under which plasticity loss occurs. Evidence suggests that mitigating plasticity loss also alleviates classical RL challenges: for instance, applying LayerNorm and weight decay stabilizes Baird's counterexample~\citep{PQNParallelisedQNetwork}, and feature normalization reduces overestimation bias in continuous control~\citep{OverestimationOverfittingPlasticity}. The line of work on plasticity loss addresses two central questions:
\begin{itemize}
    \item \emph{Why do the neural networks of deep RL agents lose their learning ability~\citep{UnderstandingPlasticity, DisentanglingCausesPlasticity, AddressingPlasticityCatastrophicForgetting, PrimacyBias, ReDoDormantNeurons, OverestimationOverfittingPlasticity}?}
    \item \emph{How can the ability to learn be maintained~\citep{L2Init, PLASTIC, HareTortoiseNetworks}?}
\end{itemize}
\textbf{These questions matter beyond deep RL}: any setting requiring adaptation to changing circumstances faces similar challenges, including continual learning~\citep{AddressingPlasticityCatastrophicForgetting} and the ubiquitous pre-train/fine-tune paradigm in supervised learning~\citep{berariuStudyPlasticityNeural2021, HareTortoiseNetworks}.

\paragraph{Contributions} This survey makes four contributions. First, we propose a unified definition of plasticity loss that subsumes prior formalizations as special cases (Section~\ref{sec:preliminaries}). Second, we provide the first systematic taxonomy of mechanisms, pathologies, and mitigation strategies. We cover approximately 50 methods organized by mechanism in Sections~\ref{sec:causes} and~\ref{sec:mitigation}, summarized in Figure~\ref{fig:plasticity-bipartite}. Third, we identify a recurring empirical pattern by demonstrating that general-purpose regularization techniques from supervised learning consistently outperform domain-specific interventions specifically designed for plasticity preservation. Fourth, we synthesize open problems and methodological gaps, providing concrete recommendations for future research (Section~\ref{sec:discussion_directions}).

\paragraph{Scope.} This survey focuses on plasticity loss in deep RL, with only brief discussions of related phenomena in continual learning and supervised learning. Existing continual learning surveys~\citep{ContinualLearningSurvey} address plasticity loss only in conjunction with catastrophic forgetting and do not focus on RL as a primary domain. Consequently, they examine a fundamentally different set of methods and lack the depth necessary for a thorough understanding of plasticity loss in deep RL. Our work also differs from \citet{ContinualRLSurvey}'s survey on continual RL, which addresses broader topics such as credit assignment and skill learning. We emphasize connections between plasticity loss and other deep RL challenges, including overestimation bias~\citep{OverestimationOverfittingPlasticity} and scaling failures~\citep{StopRegressing}. Within deep RL, we concentrate on the single-agent setting, where the understanding of plasticity loss is most developed.

\paragraph{Structure.} Section~\ref{sec:preliminaries} introduces notation, the RL formalism, and our unified definition of plasticity loss. Section~\ref{sec:related_work} positions our work relative to continual learning and continual RL. Section~\ref{sec:causes} categorizes potential mechanisms and symptoms of plasticity loss, and Section~\ref{sec:mitigation} presents a taxonomy of mitigation strategies. We conclude with a discussion of open problems and future directions in Section~\ref{sec:discussion_directions}.

%% file: sections/2preliminaries.tex
\section{Notation and Preliminaries}\label{sec:preliminaries}

We now introduce our notation, briefly outline the basics of RL, and present key RL quantities relevant to plasticity loss, along with our unifying definition of the latter. Finally, Section~\ref{subsec:preliminaries:plasticity_benchmarks} reviews benchmarks used to study plasticity loss.

\subsection{General Notation}\label{subsec:preliminaries:general_notation}

We adopt the following notation:
We use lower-case bold symbols for vectors, e.g., $\vecsample \in \samplespace \subseteq \R^{d'}$ to denote an input sample from the data space $\samplespace$ of dimension $d'$. Upper-case bold symbols denote matrices, e.g., $\designmatrix \in \R^{n \times d'}$ denotes the design matrix whose rows contain samples from $\samplespace$.
Expectations with respect to a distribution $P$ are denoted as $\E_{\mathbf{a} \sim P}[\cdot]$.
If it is clear from the context, we skip the subscript for brevity. We use $\text{SVD}(\mathbf{A})$ to denote the multiset of all singular values of $\mathbf{A}$, $\sigma$ to denote a single singular value, $\sigma_i(\mathbf{A})$ to denote the $i$th largest singular value of matrix $\mathbf{A}$ and $\sigma_{\text{min}}$ and $\sigma_{\text{max}}$ to denote the smallest and largest singular value, respectively. 
For $\representation\colon \mathbb{R}^{d'} \rightarrow \mathbb{R}^d$ being a function mapping samples to features, we denote the feature matrix as $ \representation (\designmatrix) \in \R^{n \times d}$, where $d$ is the dimension of the representation.

\subsection{Reinforcement Learning}\label{subsec:preliminaries:rl}

In RL, the goal is to optimize the reward received from an environment after performing an action. This interactive process is formalized via Markov Decision Processes (MDPs) described by tuples $\mathcal{M} = (\mathcal{S}, \mathcal{A}, \mathcal{P}, r, \rho_0, \gamma )$,
where $\mathcal{S}$ is the state space, 
$\mathcal{A}$ is the set of possible actions (action space), 
$\mathcal{P}\colon \mathcal{S} \times \mathcal{S} \times \mathcal{A} \rightarrow [0,1]$ a transition kernel specifying the probability of transitioning from one state to another upon taking a specific action, 
$r\colon \mathcal{S} \times \mathcal{A} \rightarrow \mathbb{R}$ is the reward function specifying the reward the agent obtains for taking an action in a state, 
$\rho_0$ is the initial state distribution, and 
$\gamma$ is the so-called discount factor.
The possibly stochastic policy $\pi\colon \mathcal{S} \rightarrow [0,1]^{|\mathcal{A}|}$ specifies for each state a distribution over the actions and thus determines the agent's behavior. We often write $\pi(a | s)$ to denote the probability of action $a$ in state $s$ according to policy $\pi$.
The agent aims to maximize the (discounted) cumulative reward
\begin{align}
    J(\pi) = \mathbb{E} \left[ \sum_{t=0}^\infty \gamma^t r(s_t, a_t) \right],
\end{align}
where actions are taken according to the agent's policy $\pi$ and the expectation is over the randomness of the transitions, the agent's policy, and the initial state. 
An optimal policy $\pi^*$ maximizes $J(\pi)$.
Key quantities for RL algorithms are the \emph{state-value},
\begin{align}
    V^\pi(s) = \mathbb{E}\left[ \sum_{t=0}^\infty \gamma^t r(s_t,a_t) \mid s_0 = s\right],
\end{align}
i.e., the expected cumulative reward when starting from state $s$ and following policy $\pi$ from there, and the \emph{action-value},
\begin{align}
  Q^\pi(s,a) = \mathbb{E}\left[ \sum_{t=0}^\infty \gamma^t r(s_t,a_t) \mid s_0 = s, a_0=a\right],
\end{align}
i.e., the expected return starting from state $s$, taking action $a$, and following policy $\pi$ afterwards.
An optimal policy can be found by maximizing the expected value of the initial state, i.e.,
\begin{align}
    \pi^* \in \arg \max_\pi \E_{s \sim \rho_0}[V^\pi(s)]
\end{align}
Note that state-values (and, similarly, action-values) can also be defined recursively:
\begin{align}
    V^\pi(s) &= \E_{a \sim \pi(s)}[r(s,a) + \gamma \sum_{s'} \mathcal{P}(s,s',a) V^\pi(s')], %
\end{align}
Inspired by these recursive definitions are so-called \emph{Temporal-Difference} (TD)learning approaches, e.g., approaches based on iteratively updating state-value estimates as
\begin{align}
    \hat{V}^\pi(s_t) \leftarrow \hat{V}^\pi(s_t) + \alpha [ \underbrace{r_{t+1} +\gamma \hat{V}^\pi(s_{t+1}) - \hat{V}^\pi(s_t)}_{\text{TD error}}]~.
\end{align}
In short, estimates of state- or action-values are updated based on estimates of future states (and actions), which is why such methods are also called \emph{bootstrapping methods}. The term within brackets is also referred to as \emph{TD error}.

In deep RL agents, $V^\pi,Q^\pi$ or $\pi$ (or combinations of those) are represented by deep neural networks.
Many works~\citep{AdaptiveRegularizationofRank, LyleAuxiliaryTasksRepresentation} decompose a deep RL agent into a learned representation $\representation$, covering all layers up to and including the penultimate layer, and a linear transformation $\weightmatrix$. This allows viewing an RL agent's policy or value function as a linear function of some learned non-linear features obtained through a non-linear transformation of the states $\representation (s)$ or corresponding observations. Using the value function as an example, our notation for this decomposition is $V (s) = \langle \representation (s), \weightmatrix \rangle$.

\subsection{Definition of Plasticity Loss}\label{subsec:preliminaries:definitions_plasticity}

In the literature, plasticity loss lacks a unified definition. Here, we consolidate existing definitions and demonstrate that many prior ones arise as special cases of our formulation.
Intuitively, all aim to quantify a model’s diminished ability to fit new targets but differ in how they formalize this and their training-evaluation setup. Our unified definition reads:
\begin{definition}[Loss of plasticity]\label{def:plasticity_loss}
Let $P^{(t)}_{\samplespace,\targetspace}$ be a distribution over inputs in $\samplespace$ and targets in $\targetspace$, and let $\loss^{(1)}, \loss^{(2)}, \ldots$ be a sequence of real-valued loss functions with domain $\samplespace \times \targetspace$. Let $g_\params$ represent a neural network with parameters $\params$, $\mathcal{O}$ correspond to an optimization algorithm, potentially with an optimization budget, and $\mathcal{I}$ represent an \emph{intervention} on the parameters $\params$. We denote the loss of the neural network at time $t$ using parameters $\params$ as
\begin{align}
    c^{(t)}(\params) = \mathbb{E}_{(\vecsample,\vectarget)  \sim P_{\samplespace,\targetspace}^{(t)}}\left[\loss^{(t)}(g_{\params}(\vecsample), \vectarget) \right].
\end{align}
Based on this, we define the \emph{loss of plasticity} as
\begin{align}
    &\mathcal{C}(\{ P_{\samplespace,\targetspace}^{(t)} \}_{t=1}^T, \{ L^{(t)} \}_{t=1}^T, \mathcal{O}, \mathcal{I}) = c^{(T)}(\mathcal{O}(\params^{(T)'}, P_{\samplespace,\targetspace}^{(T)}, L^{(T)})) - c^{(T)}(\tilde{\params}^{(T)})
\end{align}
where
\begin{align}
    \params^{{(t+1)}}=\mathcal{O}\left(\params^{(t)'}, P_{\samplespace,\targetspace}^{(t)}, L^{(t)} \right), \quad \params^{(t)'} = \mathcal{I}(\params^{(t)}, P_{\samplespace,\targetspace}^{(t)}, L^{(t)}), \quad \text{and} \quad \tilde{\params}^{(t)} = \mathcal{O}\left(\params^{\text{init}}, P_{\samplespace,\targetspace}^{(t)}, L^{(t)} \right).
\end{align}
\end{definition}
Here $\params^{\text{init}}$ denotes random initial parameters. 

This definition generalizes many existing definitions in that it enables different losses at different time steps, which is, e.g., relevant for multi-task learning, and in that it allows for explicit manipulations of the parameters outside of the behavior of the optimization algorithm.\footnote{Interventions on the parameters, e.g., resetting parameters to revive dead neurons, could also be considered as part of the optimizer. However, making the interventions explicit and not considering them as part of the optimizer can help clarify the different mechanisms that affect loss of plasticity.} Note that our definition focuses solely on final-task performance. This is in contrast with metrics commonly used in continual learning, such as average accuracy~\citep{ContinualLearningSurvey}, which aggregate performance over all tasks.

Many definitions of \emph{loss of plasticity} in the literature are special cases of the above definition, though some authors refer to this phenomenon by different terms:
\begin{itemize}
    \item \citet{berariuStudyPlasticityNeural2021} defined the \emph{generalization gap} as \quoteeng{the difference in performance between a pretrained model (e.g., one that has learned a few tasks already) versus a freshly initialized one}. The notion of the already-learned tasks corresponds to different tasks given by $P_{\samplespace,\targetspace}^{(t)}, \loss^{(t)}$ for $t=1, \ldots, T-1$ while the performance is evaluated with respect to a final task characterized by $P_{\samplespace,\targetspace}^{(T)}, \loss^{(T)}$. The freshly initialized model is given by $g_{\params^{(0)}}$ with $\params^{(0)}$ being random initial parameters.

    \item \citet{InFeRUnderstandingCapacityLoss} defined the \emph{target-fitting capacity} as a measure of how well a neural network can fit a distribution of targets given by a family of labeling functions (real-valued functions mapping inputs from $\samplespace$ to targets). This definition arises from Definition~\ref{def:plasticity_loss} by considering $T=1$ and selecting $P_{\samplespace,\targetspace}^{(t)}, \loss^{(t)}$ accordingly.\footnote{There is still a slight difference between the definition in~\citet{InFeRUnderstandingCapacityLoss} and our definition: our definition subtracts $c^{(T)}(\params^{(0)})$ as a baseline.} 

    \item \citet{UnderstandingPlasticity} also define \emph{loss of plasticity} but do not explicitly account for time-dependent distributions $P_{\samplespace,\targetspace}^{(t)}, \loss^{(t)}$ and interventions. Their definition is thus a special case, where no intervention is applied and constant distributions are used for both the input and the loss functions.

    \item \citet{AddressingPlasticityCatastrophicForgetting} provide a sample-based notion of plasticity loss corresponding to a baseline normalized version of the plasticity loss defined in~\citet{UnderstandingPlasticity}.
    Their definition arises as a special case of ours by fixing the loss function (i.e., using the same loss functions for all $t$) while making it dependent on the optimizer and the intervention.
\end{itemize}

\subsection{Common Benchmarks in Deep Reinforcement Learning and Plasticity Loss}\label{subsec:preliminaries:plasticity_benchmarks}

Plasticity loss can arise naturally during learning or be artificially induced for study by artificially injecting non-stationarity into a stationary learning problem. Accordingly, benchmarks can be categorized into two types: RL environments with inherent non-stationarity and supervised learning datasets with artificially introduced non-stationarity. 

\paragraph{RL Benchmarks.} Table~\ref{tab:preliminaries:rl_benchmarks} lists RL benchmarks for plasticity loss. The most well-established are Atari~\citep{Atari} (discrete actions, image observations) and DeepMind Control Suite (DMC)~\citep{DMControlSuite} (continuous actions, image or vector observations). Both benchmarks contain diverse sets of environments, including ones where plasticity loss occurs strongly. For Atari,  different game subsets have been identified which exhibit plasticity loss, with commonly studied examples including Phoenix, Space Invaders, Seaquest, Demon Attack, and Asterix~\citep{PlasticityInjection, ReDoDormantNeurons, AdaptiveRationalActivations}. For DMC, \citet{OverestimationOverfittingPlasticity} identify the Dog environment as particularly challenging due to exploding gradients during training. Additional benchmarks include Atari-100k~\citep{Atari100kSimPLe}, a 26-game subset that exacerbates plasticity loss through high replay ratios (many gradient updates per environment step)~\citep{PrimacyBias, ReplayRatioBarrier}, and MuJoCo~\citep{MuJoCo}, which has largely been superseded by DMC.

\begin{table}[h]
    \centering
    \caption{\textbf{Deep RL benchmarks for plasticity loss.}}
    \label{tab:preliminaries:rl_benchmarks}
    \footnotesize
    \begin{tabular}{@{}p{10.5cm} p{3cm}@{}}
    \toprule
    \textbf{Benchmark} & \textbf{Introduced by} \\
    \midrule
    Non-stationary MuJoCo~\citep{MuJoCo}  & \citet{ContinualBackprop}\\
    DeepMind Control Dog~\citep{DMControlSuite}  & \citet{OverestimationOverfittingPlasticity} \\
    \textbf{Atari~\citep{Atari} subset}: Phoenix, Yars revenge, Surround, Seaquest, Alien, Enduro, Asteroids, Gopher  & \citet{PlasticityInjection} \\
    \textbf{Atari~\citep{Atari} subset}: Demon attack, Asterix  & \citet{ReDoDormantNeurons} \\
    \textbf{Atari~\citep{Atari} subset}: Asterix, Enduro, Qbert, Jamesbond, Seaquest, Time pilot & \citet{AdaptiveRationalActivations} \\
    Atari-100k~\citep{Atari100kSimPLe}  & \citet{PrimacyBias} \\
    \bottomrule
    \end{tabular}
\end{table}

\paragraph{Synthetic Benchmarks.} Table~\ref{tab:preliminaries:other_benchmarks} shows synthetic benchmarks that artificially induce non-stationarity in supervised datasets. Common approaches include: 
\begin{enumerate*}[label=(\arabic*)]
    \item \emph{input shifts} via pixel shuffling~\citep{ContinualBackprop}),
    \item \emph{label shifts} through consistent permutation~\citep{UnderstandingPlasticity} or random noise~\citep{ITERNonstationarity, HareTortoiseNetworks}, and
    \item \emph{data scarcity} through reduced dataset size~\citep{HareTortoiseNetworks}.
\end{enumerate*} 
These modifications have been applied to MNIST, CIFAR-10, and ImageNet. Continual ImageNet~\citep{ContinualBackprop} is a complementary benchmark using sequential binary classification tasks drawn from ImageNet's 1000 classes. All of these modifications can also be applied in \emph{warm-starting} settings, where a network is pre-trained with non-stationary data and then fine-tuned on clean data. This setup mirrors the ubiquitous fine-tuning paradigm for pre-trained models and tests whether early training conditions degrade plasticity~\citep{berariuStudyPlasticityNeural2021, HareTortoiseNetworks, WeightClipDeepRL, DisentanglingCausesPlasticity}. Lastly, \citet{DisentanglingCausesPlasticity} note that value-based RL uses regression rather than classification~\citep{DQNOriginal, SACHaarnojaOriginal}, and develop a regression benchmark with oscillating large-mean targets, better reflecting RL optimization challenges.

\begin{table}
    \centering
    \caption{\textbf{Synthetic benchmarks for plasticity loss.}}
    \label{tab:preliminaries:other_benchmarks}
    \footnotesize
    \begin{tabular}{@{}p{2.7cm} p{5.5cm} p{5.5cm}@{}}
    \toprule
    \textbf{Benchmark} & \textbf{Non-stationarity} & \textbf{Introduced by} \\
    \midrule
     Non-stationary MNIST  & Pixel shuffling, Label shuffling, Label noise, Small data & \citet{PermutedMNIST}, \citet{UnderstandingPlasticity},\citet{HareTortoiseNetworks} \\
     Non-stationary CIFAR10 & Label noise, Label shuffling, Small Data & \citet{ITERNonstationarity}  \\
     Non-stationary ImageNet & Classification sequence, Label shuffling, Label noise, Small data  & \citet{ContinualBackprop}, \citet{AddressingPlasticityCatastrophicForgetting}, \citet{HareTortoiseNetworks} \\
     Large target regression & High-frequency targets & \citet{DisentanglingCausesPlasticity} \\
    \bottomrule  
    \end{tabular}
\end{table}

%% file: sections/3related_work.tex
\section{Related Work}\label{sec:related_work}

Plasticity loss in deep RL overlaps with two closely related fields: continual learning and continual RL. Continual learning addresses scenarios where models learn incrementally from changing data distributions or tasks, typically emphasizing the stability-plasticity trade-off to mitigate catastrophic forgetting~\citep{ContinualLearningSurvey}. Recent comprehensive surveys outline theoretical foundations, categorize methodological approaches (e.g., regularization-based, replay-based, architecture-based), and highlight practical applications across varied domains~\citep{ContinualLearningSurvey, ContinualLearningNNReview}. Specifically, the adaptation and continual fine-tuning of large language models pose advanced challenges of continual learning, requiring adaptation from general to specific capabilities as well as adaptation across time~\citep{ContinualLearningLLMsSurvey}.

In contrast, this survey specifically focuses on plasticity loss within deep RL, where shifts in the data distribution naturally arise without explicit task boundaries. Such non-stationarity arises inherently due to policy updates or improved value estimates, which affect the agent's input and target distributions. Unlike continual learning, catastrophic forgetting is not central here; rather, the emphasis is exclusively on a network's capacity to continually update parameters in response to evolving learning signals~\citep{PlasticityInjection, ResettingAdamMoments, DisentanglingCausesPlasticity}. Thus, even stationary environments can exhibit significant plasticity loss, distinguishing deep RL from classical continual learning.

Continual RL addresses non-stationarity from internal policy changes and external dynamics in transition or reward functions. This setting introduces challenges such as exploration, credit assignment, and goal-conditioned learning~\citep{ContinualRLSurvey, ContinualRLDefinition}. Our scope, however, specifically targets plasticity loss in deep RL. We focus on the agent's diminished ability to train and adapt, excluding broader issues integral to continual RL.

%% file: sections/4causes.tex
\section{Drivers and Pathologies of Plasticity Loss}\label{sec:causes}

\begin{figure}
\centering
    \includegraphics[width=\textwidth]{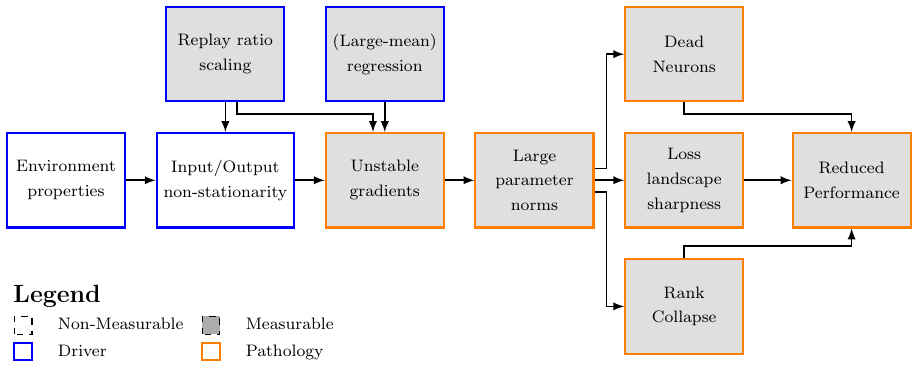}
    \Description{Flowchart with drivers and pathologies of plasticity loss.}
    \caption{\textbf{Possible connections between drivers and pathologies of plasticity loss in value-based RL}. Large-mean regression targets, combined with the non-stationarity of deep RL training, cause large and unstable gradients, leading to an increase in parameter norms. Large parameter norms are known to increase loss landscape sharpness and cause other pathologies, together leading to reduced agent performance.}
    \label{fig:plasticity_loss_model}
\end{figure}

This section categorizes the contributing factors of plasticity loss identified in the literature. As visualized in Figure~\ref{fig:plasticity_loss_model}, we distinguish between \textcolor{blue}{\textbf{drivers}} and \textcolor{orange}{\textbf{pathologies}}. Drivers encompass high-level properties of the learning problem or algorithmic choices that induce plasticity loss, such as environment properties (Section~\ref{subsec:causes:env_properties}), non-stationarity (Section~\ref{subsec:causes:nonstationarity}), high replay ratios (Section~\ref{subsec:causes:high_rr}), and the objective function (Section~\ref{subsec:causes:objective_function}). Pathologies describe the resulting internal network states or optimization dynamics associated with lost plasticity. We review these downstream effects, including parameter norm growth (Section~\ref{subsec:causes:parameter_norm_growth}), representation rank collapse (Section~\ref{subsec:causes:rank_collapse}), and saturated units (Section~\ref{subsec:causes:dead_neurons}). Finally, we discuss optimization-specific pathologies, such as gradient instability (Section~\ref{subsec:causes:first_order_effects}), loss landscape curvature (Section~\ref{subsec:causes:curvature}), and early overfitting (Section~\ref{subsec:causes:early_overfitting}).

\subsection{Environment Properties}\label{subsec:causes:env_properties}

Deep RL benchmarks exhibit substantial variation in plasticity loss, suggesting that environment-specific properties are causal factors. An example of this is changing environment dynamics over time. Atari can be categorized into stationary, dynamic, or progressive games, depending on how their input distributions evolve~\citep{AdaptiveRationalActivations}. \quoteeng{Stationary} games change little or not at all, \quoteeng{dynamic} games shift independently of the agent (e.g., new gear in \textsc{Asterix}), and \quoteeng{progressive} games adapt to the agent’s progress, as in \textsc{JamesBond} and \textsc{Montezuma’s Revenge}. Another example of changing environment dynamics is procedurally generated benchmarks such as ProcGen, which introduce variability by individually generating each level~\citep{ProcGen}. Notably, these changes do not affect all algorithms equally: off-policy methods like DQN, which store past transitions in a replay buffer, are typically more sensitive than on-policy methods such as PPO~\citep{StudyOffPolicyProceduralGeneration}.

Target non-stationarity induced by the environment may also play a role: it can increase gradient norms and reduce plasticity~\citep{DisentanglingCausesPlasticity,BRO}. In Atari’s \textsc{Seaquest}, for instance, reward magnitudes grow as the game progresses~\citep{Atari}. Reward clipping mitigates this growth but potentially discards useful signals~\citep{DQNOriginal}. \textsc{Dog} and \textsc{Humanoid} environments of the DeepMind Control Suite~\citep{DMControlSuite} are continuous-control tasks with high-dimensional action spaces, which adds a different layer of difficulty. These environments cause plasticity loss by generating large, divergent gradients~\citep{OverestimationOverfittingPlasticity,BRO}. While regularization can mitigate large-magnitude gradients, it is not guaranteed that regularization strategies succeeding in one suite (e.g., DM Control) also succeed in others (e.g., MetaWorld)~\citep{OverestimationOverfittingPlasticity}.

\subsection{Non-stationarity}\label{subsec:causes:nonstationarity}

Non-stationarity is simultaneously the most likely culprit for causing plasticity loss and one of the most elusive~\citep{ITERNonstationarity, InFeRUnderstandingCapacityLoss, PLASTIC, ContinualBackprop, HareTortoiseNetworks}. However, determining precisely how and why it affects learning remains challenging, mainly due to difficulties quantifying the degree of non-stationarity. In the following, we examine different aspects of non-stationarity that may contribute to plasticity loss.

\textbf{Input Non-stationarity} is defined through changes in the input data distribution $P(\vecsample)$~\citep{PLASTIC}. In deep RL, this often occurs due to a changing policy that generates data from a slightly different distribution with each update. Input distribution shifts may require neural networks to re-learn representations entirely, a task at which they typically struggle due to limited plasticity~\citep{AddressingPlasticityCatastrophicForgetting}. This mismatch between the need to re-learn under distribution shifts and the network's limited adaptive capacity can contribute to plasticity loss~\citep{AddressingPlasticityCatastrophicForgetting}. In experiments, such shifts can be induced by permuting input pixels or by progressively expanding datasets~\citep{ITERNonstationarity, PLASTIC, HareTortoiseNetworks, DisentanglingCausesPlasticity} (cf. Section~\ref{subsec:preliminaries:plasticity_benchmarks}). The severity of performance degradation typically correlates with the extent of the distribution shift. 
When initially learning on smaller subsets of data, these shifts can cause networks to overfit, which damages subsequent learning and generalization. This damage persists despite later exposure to a complete dataset~\citep{ITERNonstationarity, DisentanglingCausesPlasticity, HareTortoiseNetworks}.

\textbf{Target Non-stationarity} occurs when the label distribution $P(y|\vecsample)$ changes~\citep{PLASTIC}. In deep RL, this naturally occurs with temporal difference learning due to bootstrapping (cf. Section~\ref{subsec:preliminaries:rl}). Multiple studies indicate that target non-stationarity driven by bootstrapping significantly contributes to plasticity loss by causing issues such as representation collapse~\citep{FeatureRankKumar}, growing parameter norms~\citep{DisentanglingCausesPlasticity}, early performance collapse~\citep{RevisitingPlasticityAugmentations}, or dormant neurons~\citep{ReDoDormantNeurons}. All these pathologies are associated with degraded learning performance. However, the literature is conflicting on the precise role of target non-stationarity. \citet{ReDoDormantNeurons} observed dormant neurons even in offline RL settings, suggesting target non-stationarity has a more prominent role than input non-stationarity in plasticity loss. In contrast, \citet{AddressingPlasticityCatastrophicForgetting} argue that target non-stationarity relates more closely to catastrophic forgetting than plasticity loss, as adjusting to label changes theoretically requires updating only the output layer rather than fully re-learning representations.

\textbf{Trainability and Generalizability} are two distinct yet interconnected aspects of network plasticity loss~\citep{HareTortoiseNetworks, studyPlasticityOnPolicyRL}. \citet{HareTortoiseNetworks} examine fine-tuning of pre-trained models (warm-starting) under non-stationarity. They find that while warm-started models maintain high training accuracy, the non-stationarity affects generalizability as measured by test error. However, the authors are not able to establish a definitive mechanism linking non-stationarity directly to plasticity loss~\citep{HareTortoiseNetworks}. In another study, \citet{studyPlasticityOnPolicyRL} investigate plasticity loss specifically within on-policy RL using PPO, analyzing various types of non-stationarity. Their findings indicate a clear degradation in both training and testing performance as non-stationarity increases. In contrast to \citet{HareTortoiseNetworks}, \citet{studyPlasticityOnPolicyRL} establish a link between non-stationarity and rising parameter norms, suggesting that parameter growth negatively impacts both trainability and generalizability.

In summary, there is overwhelming evidence that non-stationarity is a crucial factor in plasticity loss, although the precise mechanisms remain unclear~\citep{DisentanglingCausesPlasticity, LyleNormalizationEffectiveLR}. We believe that focusing more narrowly on specific forms of non-stationarity, such as target shifts~\citep{PLASTIC}, is promising for understanding the mechanisms linking non-stationarity and plasticity loss. Clarifying these mechanisms has the potential to advance both the theoretical foundations and practical solutions to plasticity loss in deep RL.

\subsection{High Replay Ratio Training}\label{subsec:causes:high_rr}

Optimization hyperparameters play a critical role in the training dynamics of deep RL agents and have a significant impact on the phenomenon of plasticity loss. Among these, the \emph{replay ratio} (RR), sometimes also called the update-to-data (UTD) ratio, is arguably the most important hyperparameter affecting plasticity loss.

The RR describes the number of gradient updates per environment step \citep{PrimacyBias, REDQ} for off-policy deep RL agents. For instance, the original DQN agent utilizes RR=0.25, meaning one gradient step for every four environment steps \citep{DQNOriginal}, whereas SAC employs RR=1 \citep{SACHaarnojaOriginal}. While higher RRs can offer improved sample efficiency, many algorithms do not leverage them due to observed performance degradation and the emergence of degenerate policies \citep{PrimacyBias, ReplayRatioBarrier}. This degradation has been attributed to overfitting on early samples \citep{PrimacyBias} or, more recently, early plasticity loss \citep{ReplayRatioBarrier}. \citet{RevisitingPlasticityAugmentations} hypothesize that elevated RRs exacerbate initial target non-stationarity, which in turn leads to agents losing plasticity prematurely during training. A common intervention to mitigate the adverse effects of high RR training in off-policy deep RL is the application of resets, which have demonstrated efficacy across a range of benchmarks \citep{PrimacyBias, ReplayRatioBarrier, BiggerBetterFaster, BRO, OverestimationOverfittingPlasticity}.

In our view, it is unlikely that high-replay ratio training is the main cause of plasticity loss. Plasticity loss also manifests in scenarios without high RR training, such as with a standard Double DQN agent on Atari games like Phoenix and Space Invaders \citep{PlasticityInjection}. It appears that high RRs \emph{amplify} plasticity loss rather than directly causing it, which is supported by \citet{OverestimationOverfittingPlasticity}, who note that high RRs induce distinct training dynamics compared to low RRs. A prominent example is DeepMind Control's Dog environment, where SAC agents with high RRs suffer from exploding gradients \citep{OverestimationOverfittingPlasticity}. Remedies for the amplifying effect of high RRs on plasticity loss can be found in Sections~\ref{subsec:mitigation:untargeted_resets}, \ref{subsec:mitigation:weight_regularization}, \ref{subsec:mitigation:feature_regularization}, \ref{subsec:mitigation:architectures}, \ref{subsec:mitigation:other_regularizers} and \ref{subsec:mitigation:combined_methods}.

Beyond the replay ratio, other optimization hyperparameters significantly influence plasticity. Larger learning rates can aid in escaping suboptimal local minima in supervised learning, whereas in offline deep RL, high learning rates are prone to rank collapse and dead neurons \citep{gulcehreEmpiricalStudyImplicit}. Implicit learning rate scheduling may also play a larger role than previously thought \citep{LyleNormalizationEffectiveLR}, underscoring the need for careful tuning. The total number of training steps intuitively impacts plasticity: longer training on one task can lead to reduced adaptability for subsequent tasks \citep{CurvatureExplainsPlasticity}. Finally, the batch size interacts with both gradient noise and learning rate: smaller batches can reduce gradient collinearity and sometimes improve performance~\citep{SmallBatchDeepRL}, while larger batches support stability at higher learning rates but may also increase the risk of dead units~\citep{gulcehreEmpiricalStudyImplicit}. Therefore, empirically tuning hyperparameters for a given task and environment is often necessary to achieve optimal performance and plasticity~\citep{hyperparameterconsistency}.

\subsection{Objective Function}\label{subsec:causes:objective_function}

Classification tasks have been observed to be easier for deep neural networks than regression ever since the AlexNet breakthrough in image classification \cite{StopRegressing}. Because value-based deep RL methods use a regression loss, recent work has hypothesized that regressing on non-stationary targets might be one of the leading causes of deep RL's optimization issues~\citep{StopRegressing, DisentanglingCausesPlasticity}. For example, the two-hot trick~\citep{TwoHotMuZero} has been successfully applied to train value networks and has since been linked to mitigating plasticity loss, albeit at a performance cost~\citep{UnderstandingPlasticity}. \citet{StopRegressing} find that reformulating regression as classification using a method called HL-Gauss~\citep{HL-GaussOriginal} improves RL agents in many ways, such as representational capacity, robustness to noise and uncertainty, and sample efficiency. However, they do not provide deeper insights into the exact mechanisms behind regression optimization issues.

A recent investigation into the mechanisms driving plasticity loss hypothesizes that \emph{regression with large-mean targets} causes networks to lose plasticity even in stationary learning problems~\citep{DisentanglingCausesPlasticity}. This issue is particularly prevalent in value-based deep RL, where an agent ideally improves throughout training, resulting in temporal difference targets with increasing magnitude. Regressing on large-mean targets has two negative side effects: First, deep networks are prone to encoding the target offset into their weights instead of the bias. This leads to an explosion of the singular values of the corresponding dimensions in parameter space, resulting in an ill-conditioned feature matrix~\citep{DisentanglingCausesPlasticity}. Second, when the network predicts the large-mean targets insufficiently well, the squaring in the mean squared error yields large error terms. As gradients are proportional to errors in regression tasks, these large errors potentially cause the parameter norms of the network to grow rapidly~\citep{StopRegressing, DisentanglingCausesPlasticity}. This growth of the parameter norms is associated with a wide range of pathologies such as poor generalization~\citep{WeightClipDeepRL}, loss landscape sharpness~\citep{DisentanglingCausesPlasticity}, and finally plasticity loss~\citep{DisentanglingCausesPlasticity}. Feature regularization (Section~\ref{subsec:mitigation:feature_regularization}) mitigates the downstream effects of regression losses, while Section~\ref{subsec:mitigation:loss_reformulation} covers categorical loss functions in detail.

\subsection{Parameter Norm Growth}\label{subsec:causes:parameter_norm_growth}

\citet{DisentanglingCausesPlasticity} found that parameter norm growth is a common pathology of networks that have lost plasticity and is associated with reduced task performance. In the following, we present four possible mechanisms to explain this effect.
First, \citet{DisentanglingCausesPlasticity} link growing parameter norms and the sharpness of the optimization landscape: As the norms of the parameters grow, so does the maximum eigenvalue of the Hessian. This connects to the works described in Section~\ref{subsec:causes:curvature}, hypothesizing that curvature may explain plasticity loss. Second, the same authors also hypothesize that large parameter norms may affect nonlinearities within the network, such as activation functions or softmax heads saturating. The effects of saturated units are described in detail in Section~\ref{subsec:causes:dead_neurons}, but among them are gradient propagation issues and a loss in effective network capacity, both of which are associated with plasticity loss.
Third, it has been found that growing parameter norms may lead to gradients with sparse and/or collinear gradient covariance matrices. As a result, updates may over-generalize or under-generalize in the sample space. In an extreme case, an over-generalizing network may learn a degenerate function that assigns similar outputs to all inputs. On the other hand, an under-generalizing network can only memorize task labels~\citep{DisentanglingCausesPlasticity}. Both states are linked to plasticity loss.
Fourth, a recent study finds that large parameter norms affect the effective learning rate of a network. In particular, as the parameter norms of a network grow, so does the norm of its gradient, leading to instability and potential plasticity loss during training~\citep{LyleNormalizationEffectiveLR}. The implicit effect of parameter norms on the learning rate can be mitigated with a technique called Normalize-and-Project, which we discuss in Section~\ref{subsec:mitigation:combined_methods}. Other remedies are covered in Sections~\ref{subsec:mitigation:weight_regularization}, \ref{subsec:mitigation:feature_regularization}, \ref{subsec:mitigation:loss_reformulation} and~\ref{subsec:mitigation:architectures}.

\subsection{Feature Rank Collapse}\label{subsec:causes:rank_collapse}

The effective rank is a measure commonly used to assess the quality of the representation learned by a neural network \citep{FeatureRankKumar, LyleNormalizationEffectiveLR, ContinuousFeatRank}. To build an intuition of why it matters, note that an RL agent's value function is commonly computed as $V (s) = \langle \representation (s), \weightmatrix \rangle$, i.e., as the inner product of non-linear features of the state and weights $\weightmatrix$. Now suppose that $\representation (s) \in \R^d$ is low rank: This implies that the network's features lie in a lower-dimensional subspace of $\R^d$, potentially mapping dissimilar states to similar feature vectors, which in turn makes it harder to learn distinct values for these dissimilar states \citep{StateAliasingMcCallum}. For a full-rank $\representation (s)$, the network maps different states to more dissimilar feature vectors by utilizing all directions of $\R^d$, facilitating learning of distinct values for different states.
\citet{FeatureRankKumar} define the effective rank as the minimum $k$ such that a rank-$k$ approximation of the feature matrix explains at least a $(1-\delta)$ fraction of its total variance:
\begin{definition}[Effective Rank~\citep{FeatureRankKumar}]\label{def:effective_rank}
Let $\phi\colon \samplespace \rightarrow \R^d$ be a feature mapping and $\representation (\designmatrix) \in \R^{n \times d}$ be a feature matrix, e.g., the embeddings of a neural network for a collection of samples $\designmatrix$.
Let $\delta \in [0,1]$ and $\sigma_1 \geq \cdots \geq \sigma_d \geq 0$ be the singular values of $\representation (\designmatrix)$ in decreasing order.
The \emph{effective rank} is defined as: $\operatorname{srank}_\delta(\representation (\designmatrix))=\min \left\{k: \frac{\sum_{i=1}^k \sigma_i(\representation (\designmatrix))}{\sum_{i=1}^d \sigma_i(\representation (\designmatrix))} \geq 1-\delta\right\}$.
\end{definition}
Multiple works observe a correlation between an agent's performance degradation and its representation becoming low-rank. This phenomenon is dubbed "rank collapse" and has been observed both in online~\citep{FeatureRankKumar, InFeRUnderstandingCapacityLoss, AdaptiveRegularizationofRank} and offline~\citep{FeatureRankKumar, gulcehreEmpiricalStudyImplicit} RL. \citet{FeatureRankKumar} establish a connection between the effective rank of an agent's representation and its ability to learn: A decrease in the representation's rank leads to increased TD error. In theoretical and empirical analysis, \citet{FeatureRankKumar} show that a drop in rank is associated with the largest singular values of the representation outgrowing the smaller ones, most likely caused by bootstrapping in value-based deep RL. This effect may be exacerbated in sparse-reward environments such as Montezuma's revenge on Atari~\citep{UnderstandingPlasticity}. \citet{gulcehreEmpiricalStudyImplicit} present a thorough empirical study and find that the association between effective rank and agent performance is not as straightforward as previously assumed in offline RL. In particular, it depends on potentially confounding hyperparameter and architecture choices, such as the learning rate and the activation function. However, they show that the collapse of the effective rank to a very small value is reliable and enables the identification of underfitting agents with suboptimal performance at the end of training. 

The phenomenon of rank collapse is closely tied to other phenomena related to the loss of plasticity. In offline RL, there is a strong correlation between the effective rank and the number of dead units in the agent's network~\citep{gulcehreEmpiricalStudyImplicit}. Similarly, resetting dead or dormant neurons increases feature rank at the end of online RL training~\citep{ReDoDormantNeurons}. Currently, the exact causal relationship between dead neurons and rank collapse remains unclear. Strategies to mitigate the effect of rank collapse can be found in Sections~\ref{subsec:mitigation:weight_regularization}, \ref{subsec:mitigation:feature_rank_regularization}, \ref{subsec:mitigation:activations} and~\ref{subsec:mitigation:architectures}.

\subsection{Saturated and Dormant Neurons}\label{subsec:causes:dead_neurons}

Saturated~\citep{HighVarianceRL, DisentanglingCausesPlasticity} or dormant~\citep{ReDoDormantNeurons} units are one of the most prominent pathologies associated with plasticity loss and reduced agent performance. They are an obvious and objectively measurable sign indicating that the network cannot utilize its full capacity. Therefore, it is easy to relate them to reduced network expressivity and slow learning~\citep{ReDoDormantNeurons}. However, it is unclear whether dormant neurons are a main driver of plasticity loss or just a pathology associated with networks that have lost their plasticity. For example, \citet{ReDoDormantNeurons} discuss target non-stationarity as contributing to an increase in dormant neurons throughout training. Other aspects of modern deep RL algorithms might also exacerbate the phenomenon, such as training with high replay ratios~\citep{ReplayRatioBarrier, BiggerBetterFaster}.

How can units with reduced capacity be formally defined? The literature provides a plethora of options to achieve this, which we categorize into two groups: \emph{Saturated units}, for which a shift in the pre-activation distribution reduces the neuron's capacity to produce meaningfully different outputs given inputs from its input distribution, which often coincides with vanishing gradients. For example, this could be a ReLU neuron where all inputs are positive or negative, rendering it inactive (\quoteeng{dead})~\citep{ReLUDeadNeurons} or linear~\citep{DisentanglingCausesPlasticity}, respectively. \emph{Dormant units}~\citep{ReDoDormantNeurons} are instead characterized by low post-nonlinearity activations. When considering a $\tanh$ unit, one can easily see how these categories differ: For large pre-activations, the unit's output will be close to one for all inputs. Importantly, the output will be close to one (i.e., show small numerical differences), even considering significant differences in the pre-activation. On the other hand, such a saturated $\tanh$ unit is clearly not dormant because its post-nonlinearity activation is high.

\textbf{Saturated Neurons} have first been discussed in \citet{HighVarianceRL} in the context of $\tanh$ activations in pixel-based continuous control. However, no common general definition of saturated neurons has emerged yet. The authors took a closer look at runs of the then-state-of-the-art agent DrQ-v2~\citep{DrQv2} and observed that most of them exhibited saturated $\tanh$ policies, with runs failing to learn and not being able to move out of this regime\footnote{DrQ-v2 builds on top of TD3~\citep{TD3OriginalPaper}, which is an agent based on deterministic policy gradients that uses the $\tanh$ function to clip actions in $[-1, 1]$.}. 
This pushes actions to the boundaries of the $[-1, 1]$ action interval typically used in continuous control. When $| \tanh (g_\params(\vecsample)) | \approx 1$, the derivative $1 - \tanh^2(g_\params(\vecsample)) \approx 0$, causing vanishing gradients. Interestingly, their proposed solution to normalize the features of the penultimate layer increases performance when applied not only to the actor but also to the critic~\citep{HighVarianceRL}, highlighting the fact that bounded activations may be just as important to prevent critic divergence~\citep{CrossQ}.

More recently, \citet{DisentanglingCausesPlasticity} partition non-stationary learning problems into an \emph{erasing or unlearning phase} and a \emph{disentanglement phase}, finding that the first causes a form of saturation in ReLU units. These \emph{linearized units} are characterized by their pre-activation distribution, which has only positive support. To be more precise, the unlearning phase after a task shift causes a distribution shift in the neuron's pre-activation distribution through gradients that either increase or decrease the preactivation values for all training samples. If all pre-activations for a neuron are now positive, the unit becomes \emph{linearized}, removing its nonlinear component. If a neuron's pre-activations are negative for the training dataset, it moves into the dead neuron regime. Both of the aforementioned pathologies reduce the network's expressive capacity. As a remedy, the authors propose to apply LayerNorm~\citep{LayerNorm} before a unit's nonlinearity~\citep{DisentanglingCausesPlasticity} to ensure (approximately) zero-mean pre-activations.

\textbf{Dormant Neurons} have been introduced by \citet{ReDoDormantNeurons} as a measure for the reduced expressivity of a neural network. In experiments, it has been shown that dormant neurons are associated with performance plateaus and reduced performance of DQN agents~\citep{DQNOriginal} trained on various Atari games. To determine the level of dormancy, one has to first calculate normalized activation scores $s_i^l$ for each neuron $i$ in all non-final layers $l$~\citep{ReDoDormantNeurons}, i.e.,
\begin{gather}\label{eq:dormant_neuron_scores}
s_i^l=\frac{\mathbb{E}_{\vecsample \in \buffer}\left[ |h_i^l(\vecsample) | \right]}{\frac{1}{H^l} \sum_{k \in [H^l]} \mathbb{E}_{\vecsample \in \buffer}\left[ |h_k^l(\vecsample) | \right]}~.    
\end{gather}
Here $\vecsample \in \buffer$ are samples drawn from an input distribution, e.g., the replay buffer in off-policy deep RL, $h_i^l(\vecsample)$ denotes the post-nonlinearity activation of a neuron in layer $l$, and $H^l$ the number of neurons in layer $l$.

\begin{definition}[Neuron Dormancy~\citep{ReDoDormantNeurons}]\label{def:dormant_fraction}
  If the score in Equation~\eqref{eq:dormant_neuron_scores} is below a threshold $\tau$, i.e., $s_i^l \leq \tau$, then neuron $i$ in layer $l$ is \emph{$\tau$-dormant}. Denoting $H_\tau^l$ as the number of dormant neurons per layer, and $N^l$ as the total number of neurons per layer, the \emph{dormancy ratio} $\beta_\tau$ is the fraction $H_\tau^l/N^l$ of all layers except the final layer, $\beta_\tau=\sum_{l \in \params} H_\tau^l / \sum_{l \in \params} N^l$~\citep{ReDoDormantNeurons, DrMDormantRatio}. An agent exhibits the \emph{dormant neuron phenomenon} if $\beta_\tau$ increases over the course of the training.
\end{definition}

The above definition is not the only sensible way to define a measure for neurons with reduced activity, but it is currently widely used in the literature~\citep{ReDoDormantNeurons, DrMDormantRatio, RevisitingPlasticityAugmentations, SmallBatchDeepRL, prunedNetworksAreGoodNetworks} due to its simplicity and intuitive understanding. \citet{ContinualBackprop} define a more complex measure of neuron utility based on a product between the magnitudes of a unit's summed weights and its activations. A more straightforward option for ReLU networks is to simply count the number of zero activations~\citep{AddressingPlasticityCatastrophicForgetting, CReLU}.

To summarize: While there are different options for defining inactive neurons~\citep{ReDoDormantNeurons, AddressingPlasticityCatastrophicForgetting, ContinualBackprop}, all of the works cited above agree that they are a symptom of reduced expressivity caused by some form of non-stationarity~\citep{ReDoDormantNeurons, RevisitingPlasticityAugmentations, CReLU}. Where they differ is in the proposed solution to address inactive units: Sections~\ref{subsec:mitigation:untargeted_resets} and~\ref{subsec:mitigation:targeted_resets} discuss a plethora of reset strategies as possible remedies~\citep{PrimacyBias, ReDoDormantNeurons, DrMDormantRatio, ContinualBackprop}, Section~\ref{subsec:mitigation:activations} considers activation functions for non-stationary problems such as CReLU~\citep{CReLU}, and Section~\ref{subsec:mitigation:weight_regularization} examines parameter regularization~\citep{CurvatureExplainsPlasticity}. More strategies to counter the effects of saturated or dormant neurons are covered in Sections~\ref{subsec:mitigation:feature_regularization}, \ref{subsec:mitigation:optimization}, \ref{subsec:mitigation:architectures}, \ref{subsec:mitigation:other_regularizers}, and \ref{subsec:mitigation:combined_methods}.

\subsection{First-order Optimization Effects}\label{subsec:causes:first_order_effects}

Since deep networks rely on gradient-based optimization, gradient pathologies such as the well-known \textbf{vanishing gradient problem}~\citep{HochreiterVanishingGrads} are natural candidates for plasticity loss. They manifest as a collapse in nonzero gradients and L1 gradient norms\footnote{\citet{CReLU} note that L2 gradient norms can be misleading due to disproportionate weight from outliers.}, where the network cannot adapt despite high loss. Gradient sparsity, a related phenomenon, correlates with sparse ReLU activations and dead neurons in value-based agents\citep{CReLU}.  A more nuanced notion of sparsity, \emph{gradient dormancy}, adapts the neuron dormancy measure from Section~\ref{subsec:causes:dead_neurons} by using gradient L2 norms instead of activations. \citet{ACEOffPolicyActorCriticCausallyAware} find that gradient dormancy afflicts proprioceptive control agents in sparse-reward tasks, indicating a connection between sparse feedback, degraded learning ability, and representation quality~\citep{LyleAuxiliaryTasksRepresentation}.

\textbf{Exploding gradients} destabilize training and typically cause more severe performance degradation than vanishing gradients. They arise in at least three deep RL settings. First, high-dimensional continuous control environments, such as DM Control Dog (38-dimensional action space), generate large gradient norms that can destabilize training~\citep{OverestimationOverfittingPlasticity}. Second, scaling network depth in continuous control without regularization leads to gradient explosion~\citep{DeeperDeepRLSpectralNorm}. Third, off-policy agents suffer from large gradients when combining overestimation bias with updates on out-of-distribution (OOD) actions~\citep{OFNDissectingRLHighUTD}. This third mechanism operates as follows: Bootstrapping with $\argmax$ in off-policy algorithms selects overestimated OOD actions due to function approximation error~\citep{DoubleQLearning}. These OOD actions produce large TD errors with proportionally growing critic gradients due to the  regression loss (Section~\ref{subsec:causes:objective_function}). Large gradients then increase parameter norms, further amplifying overestimation and gradient magnitudes in a feedback loop. This effect intensifies when training with high replay ratios or on small, fixed batches~\citep{PrimacyBias, OFNDissectingRLHighUTD}.

In contrast to vanishing or exploding gradients, \textbf{ill-conditioned updates} are a more subtle issue related to gradients in optimization. \citet{lewandowski2024spectralregularization} examine how non-stationarity caused by growing parameter norms leads to a collapse in Jacobian ranks. This reduces the diversity of gradients, with updates only being performed along a few dimensions in parameter space. Another symptom of ill-conditioned gradients is gradient collinearity, which can be analyzed through two related metrics: the \emph{gradient covariance matrix}~\citep{UnderstandingPlasticity} and the \emph{empirical neural tangent kernel (eNTK)}~\citep{DisentanglingCausesPlasticity}. Both examine relationships between gradients of different training samples. The gradient covariance matrix uses normalized dot products (cosine similarity) between objective gradients, whereas the eNTK employs unnormalized dot products between network output gradients. The structure of these matrices provides diagnostic insights into both optimization and generalization. Positive values between sample pairs indicate that gradient updates generalize across those samples, while negative values suggest interference. A pronounced block structure in these matrices signals a sharp and unstable loss landscape. Most matrix entries being either positive or negative and of similar magnitude indicate that the network has learned a degenerate function: updates on individual samples overgeneralize to the entire input space, making it difficult to distinguish between samples. Most of the available remedies to plasticity loss can directly mitigate the gradient issues described above (cf. Sections~\ref{subsec:mitigation:untargeted_resets}, \ref{subsec:mitigation:weight_regularization}, \ref{subsec:mitigation:feature_regularization}, \ref{subsec:mitigation:feature_rank_regularization}, \ref{subsec:mitigation:activations}, \ref{subsec:mitigation:loss_reformulation}, \ref{subsec:mitigation:optimization}, \ref{subsec:mitigation:architectures} and \ref{subsec:mitigation:combined_methods}).

\subsection{Second-order Optimization Effects}\label{subsec:causes:curvature}

It is well-known from supervised learning that flat minima generalize better~\citep{HochreiterFlatMinima, SharpnessAwareMinimization}. Curvature also appears central to plasticity loss: networks that lose plasticity often show high curvature~\citep{UnderstandingPlasticity, CurvatureExplainsPlasticity, PLASTIC}. \citet{UnderstandingPlasticity} find that increased curvature, as measured by the largest eigenvalue of a network's Hessian, makes optimization more difficult and leads to plasticity loss. In a similar vein, \citet{CurvatureExplainsPlasticity} argue that a collapse of the Hessian's effective rank~\citep{FeatureRankKumar} (cf.\ Definition~\ref{def:effective_rank}) is causing plasticity loss. To measure curvature, they find that the empirical Fisher information matrix outperforms other approximations, such as the Gauss-Newton approximation, in terms of accuracy~\citep{CurvatureExplainsPlasticity}. If a sharp optimization landscape were to cause plasticity loss, then methods explicitly reducing sharpness, such as the SAM optimizer~\citep{SharpnessAwareMinimization}, should also mitigate it (cf Section~\ref{subsec:mitigation:optimization}). \citet{PLASTIC} evaluate SAM in an Atari-100k agent and find it very effective at reducing sharpness as measured by the Hessian's largest eigenvalue. However, their ablation studies highlight that resets~\citep{PrimacyBias} (cf.\ Section~\ref{subsec:mitigation:untargeted_resets}) still outperform SAM in terms of cumulative reward when applied in isolation. The performance gap between resets and SAM occurs despite significantly higher curvature when applying the former, indicating that sharpness cannot be the sole mechanism behind plasticity loss. \citet{PLASTIC} attribute this performance gap to smooth minima only reducing sensitivity to changes in the input distribution, whereas complementary methods must tackle the orthogonal issue of changes in the target distribution. Moreover, combined strategies (Section~\ref{subsec:mitigation:combined_methods}) and parameter regularization (Section~\ref{subsec:mitigation:weight_regularization}) are useful in mitigating issues with the loss landscape.

\subsection{Primacy Bias}\label{subsec:causes:early_overfitting}

When training networks under non-stationarity, some early works have hypothesized that residual effects of overfitting to early training data might hinder late training progress. The earliest work describing this phenomenon attributes it to the network trying to re-use suboptimal features acquired early during training~\citep{ITERNonstationarity}. Their results on toy datasets have been confirmed in deep RL and subsequently named the \quoteeng{\emph{Primacy bias}}~\citep{PrimacyBias}, borrowing terminology from a psychological phenomenon where early experiences disproportionately shape later learning. In deep RL, this manifests as agents overfitting to early interactions and losing the ability to update their networks when faced with new data. While these early findings regarding plasticity loss have been compelling, more recent work has moved away from the hypothesis of early overfitting towards mechanistic and measurable explanations such as dead neurons, parameter norm growth, or feature rank collapse. Mitigation strategies to primacy bias comprise non-targeted weight resets and combined approaches (cf. Sections~\ref{subsec:mitigation:untargeted_resets},~\ref{subsec:mitigation:feature_regularization}, and \ref{subsec:mitigation:combined_methods}).

%% file: sections/5mitigation.tex
\section{Mitigating Loss of Plasticity}
\label{sec:mitigation}

\begin{figure}[t]
\centering
\includegraphics[width=\textwidth]{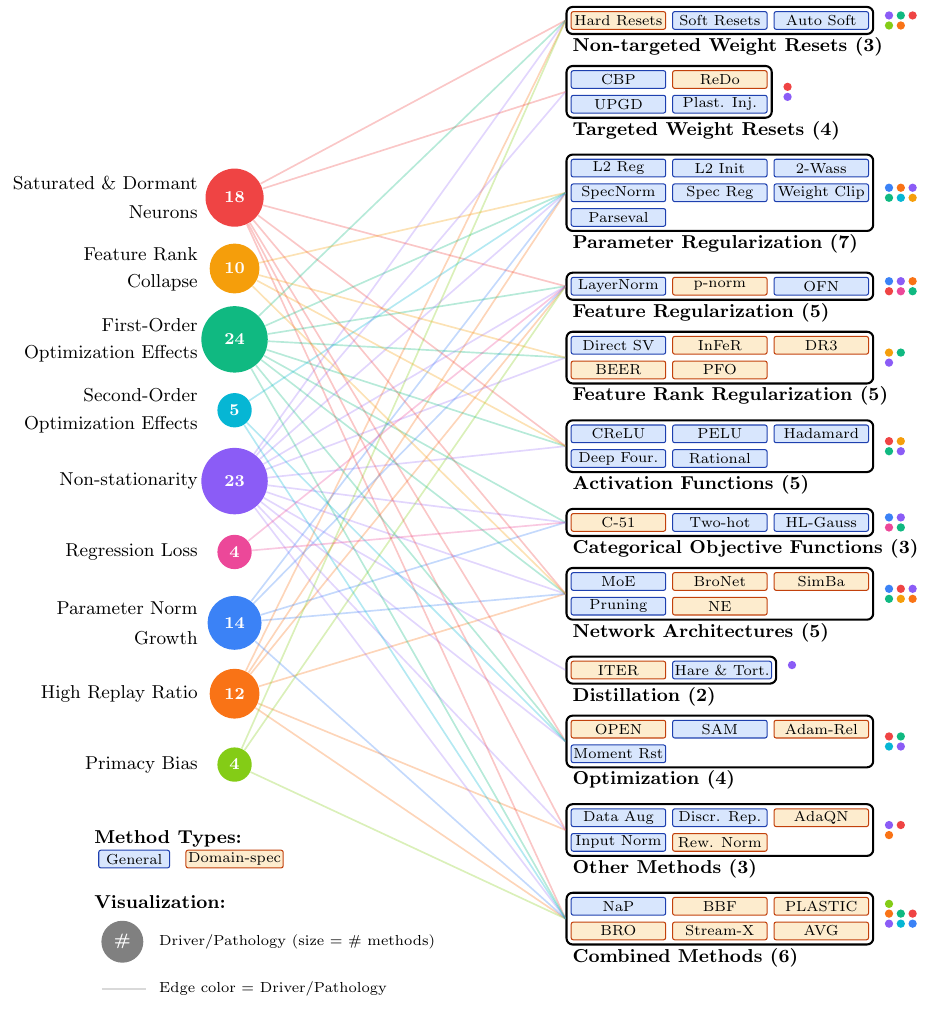}
\Description{Biparte graph with drivers and pathologies of plasticity loss as well as mitigation methods.}
\caption{Bipartite graph showing plasticity loss mitigation methods grouped by category. \textbf{Left}: Nine causes sized by the number of methods addressing them. \textbf{Right}: Methods grouped into 12 categories with individual methods shown within each category box. Edge color indicates which driver or pathology a category addresses. Dots next to each category box indicate drivers/pathologies addressed by methods within the box.}
\label{fig:plasticity-bipartite}
\end{figure}

This section provides an overview of methods for mitigating plasticity loss. We begin with network reinitialization approaches using untargeted (Section~\ref{subsec:mitigation:untargeted_resets}) and targeted (Section~\ref{subsec:mitigation:targeted_resets}) weight resets, then cover regularization of weights (Section~\ref{subsec:mitigation:weight_regularization}) and feature rank (Section~\ref{subsec:mitigation:feature_rank_regularization}). Section~\ref{subsec:mitigation:activations} examines activation functions, Section~\ref{subsec:mitigation:loss_reformulation} discusses categorical loss reformulation, and Section~\ref{subsec:mitigation:architectures} covers network architectures. We then discuss distillation-based methods (Section~\ref{subsec:mitigation:distillation}) and RL-specific optimizers (Section~\ref{subsec:mitigation:optimization}), followed by miscellaneous techniques (Section~\ref{subsec:mitigation:other_regularizers}). Section~\ref{subsec:mitigation:combined_methods} concludes with methods combining multiple mechanisms. Figure~\ref{fig:plasticity-bipartite} summarizes our taxonomy.

\clearpage

\subsection{Non-targeted Weight Resets}\label{subsec:mitigation:untargeted_resets}

Non-targeted weight resets periodically reinitialize network layers to restore plasticity and were initially proposed to mitigate early overfitting~\citep{PrimacyBias}. They come in two variants: 
\begin{enumerate*}[label=(\alph*)]
  \item \emph{Hard resets}, where network layers are fully reset, and 
  \item \emph{soft resets}, where a convex combination of current and fresh weights is used.
\end{enumerate*}
\textbf{Hard Resets} fully reset network weights using the initialization distribution. Empirical evidence shows plasticity loss concentrates in later layers~\citep{berariuStudyPlasticityNeural2021, PrimacyBias, ReplayRatioBarrier}, motivating resets of the final layers while preserving earlier representations. For shallow networks in proprioceptive control, this still amounts to resetting the entire network~\citep{PrimacyBias, ReplayRatioBarrier, BRO}. In pixel-based domains like Atari, convolutional encoders are typically preserved to retain the learned representation~\citep{PrimacyBias, ReplayRatioBarrier, BiggerBetterFaster}. \textbf{Soft Resets}~\citep{ShrinkPerturb}, a.k.a. Shrink and Perturb (S\&P), blend current weights $\params_\text{old}$ with freshly initialized parameters $\psi$ from the initial weight distribution~\citep{ReplayRatioBarrier}: $\theta_\text{new} = \alpha \theta_\text{old} + (1 - \alpha) \psi,\; \psi \sim \text{initializer}$. When $\alpha$ is well-tuned, S\&P restores plasticity while retaining task-relevant knowledge. When $\alpha$ is too high, plasticity might not be sufficiently restored; when $\alpha$ is too low, the reset can erase the learned knowledge (catastrophic forgetting).

Many agents use some form of resetting to address plasticity loss. SR-SPR~\citep{ReplayRatioBarrier} applies soft resets to the encoder with $\alpha \approx 0.8$ and hard resets to the head, enabling training with replay ratios up to 16 on Atari-100k. BBF~\citep{BiggerBetterFaster} uses the same reset strategy alongside deeper residual networks, setting the current state-of-the-art for model-free RL on Atari-100k. Several methods avoid manual $\alpha$ tuning: DrM~\citep{DrMDormantRatio} and ACE~\citep{ACEOffPolicyActorCriticCausallyAware} dynamically adjust reset strength based on plasticity metrics. DrM uses the dormancy ratio as a plasticity measure, while ACE uses gradient dormancy (see Sections~\ref{subsec:causes:dead_neurons} and~\ref{subsec:causes:first_order_effects}). \citet{nonstationaryLearningAutomaticSoftResets} take a Bayesian view on plasticity loss and model the drift of the optimal parameters due to non-stationarity with an Ornstein-Uhlenbeck process, assuming a Gaussian prior and posterior for the weights. When their drift model detects parameter drift with high probability, the algorithm adjusts the network's current weights (posterior) closer to the initialization weights (prior), enabling faster learning. Conceptually, this \textbf{Automatic Soft Parameter Reset} can be understood as a widening of the confidence region until the optimal parameters $\theta^*_t$ for the new task are contained in it with high probability \citep{nonstationaryLearningAutomaticSoftResets}, removing the $\alpha$ hyperparameter entirely.

Comparing hard and soft resets, it stands out that hard resets are primarily viable for off-policy algorithms with replay buffers, which serve as a form of "memory" that enables the rapid recovery of discarded knowledge~\citep{PrimacyBias, ReplayRatioBarrier}. Without a buffer, hard resets risk catastrophic forgetting, though on-policy use may be feasible with sufficiently low reset frequencies~\citep{studyPlasticityOnPolicyRL}. S\&P with well-tuned $\alpha$ avoids forgetting and works well in on-policy settings~\citep{studyPlasticityOnPolicyRL}. Additionally, S\&P reduces dead neurons, mitigates vanishing gradients, and controls weight norm growth~\citep{AddressingPlasticityCatastrophicForgetting}. Additionally, resets may offer exploration benefits that are orthogonal to plasticity. \citet{DrMDormantRatio} hypothesize that S\&P improves exploration by accelerating policy change~\citep{PolicyChurn}. \citet{OFNDissectingRLHighUTD} observe similar exploration benefits arising from resets that shift which actions the Q-function over- or underestimates. This suggests that resets may enable a form of optimistic exploration. In summary, hard resets offer simplicity and strong restoration of plasticity in off-policy settings, while soft resets provide finer control and broader applicability at the cost of increased hyperparameter sensitivity.

\subsection{Targeted Weight Resets}\label{subsec:mitigation:targeted_resets}

Instead of arbitrarily resetting network weights, algorithms with \emph{targeted resets} track a measure of utility for each neuron or parameter and perform resets accordingly~\citep{ReDoDormantNeurons, ContinualBackprop, AddressingPlasticityCatastrophicForgetting}. This allows resetting only the parts of a network likely affected by plasticity loss, at the cost of additional computational and memory overhead. In this section, we examine four approaches that trade computational cost for precision in identifying neurons to reset. All four methods substantially outperform non-targeted resets on tasks where plasticity loss is severe. However, their relative effectiveness varies across benchmarks and agent architectures.

\textbf{Continual Backpropagation}~\citep{ContinualBackprop} tracks a heuristic utility measure for each neuron that is updated after every gradient step. The algorithm resets the least useful $x\%$ of neurons in each layer by resampling incoming weights from the initialization distribution and setting outgoing weights to zero. It also resets the optimizer's moment estimates and maintains a counter to prevent repeatedly resetting the same neurons. While CBP shows promising results on toy problems and proprioceptive RL environments, tracking multiple metrics per neuron and performing updates at every gradient step makes it computationally expensive and memory-intensive. 
\textbf{ReDo}~\citep{ReDoDormantNeurons} reduces overhead by only checking every $k$ time steps whether a neuron's per-layer normalized activations fall below a threshold. Neurons below this threshold are considered dormant and reset using the same strategy as in Continual Backpropagation. On Atari, ReDo substantially improves DQN and DrQ performance, particularly in games where dormant neurons are known to occur. Improvements are less pronounced for SAC agents on proprioceptive continuous control tasks due to different dormancy dynamics compared to pixel-based environments~\citep{ACEOffPolicyActorCriticCausallyAware}.
\textbf{UPGD}~\citep{AddressingPlasticityCatastrophicForgetting} combines targeted resets with noisy gradient descent by scaling the learning rate with a measure of parameter utility. Parameters with high utility remain largely unchanged, while unimportant parameters receive stronger SGD updates and are perturbed with Gaussian noise. The utility approximates the change in loss if a particular parameter were set to zero, estimated via a first-order Taylor expansion to avoid expensive forward passes. This approach can be viewed as a soft, utility-weighted reset that extends naturally to momentum-based optimizers like Adam.

Unlike the previous three methods that explicitly measure neuron or parameter utility, \textbf{Plasticity Injection}~\citep{PlasticityInjection} takes a different approach by resetting the network's head in a way that preserves both outputs and trainability. Building on the observation that plasticity loss concentrates in the last layers~\citep{ReplayRatioBarrier}, it decomposes the Q-function as $Q(\vecstate) = h_{\params} (\vecstate) + h_{{\params}_1'} (\vecstate) - h_{{\params}_2'} (\vecstate)$, where $\theta$ are the original frozen parameters, $\theta'_1$ are freshly initialized trainable parameters, and $\theta'_2$ is a frozen copy of $\theta'_1$. This construction ensures that predictions remain unchanged immediately after the injection, while the fresh parameters $\theta'_1$ restore the agent's learning capacity. Because plasticity injection performs a targeted reset of the network's head without measuring individual neuron utilities, it occupies a middle ground between targeted and non-targeted resets. This dual nature also makes it useful as a diagnostic tool: if an agent's performance improves substantially after injection, this confirms that plasticity loss was hindering learning~\citep{PlasticityInjection}.

\subsection{Parameter Regularization}\label{subsec:mitigation:weight_regularization}

Regularizing parameter norms was originally developed to combat overfitting in linear regression~\citep{ProbMLIntroBook}. In contrast, avoiding plasticity loss focuses not on simplifying the model but on preserving its ability to learn. To this end, parameter regularization helps retain properties of initial weights known to support rapid adaptation, such as small parameter magnitudes, the rank of learned representations, or favorable Hessian conditioning. 

Small parameter norms may aid training by inducing a smoother optimization landscape via smaller gradient norms~\citep{DisentanglingCausesPlasticity}, making parameter regularization a natural avenue for mitigating plasticity loss. A first family of methods regularizes weights toward their initial configuration. The most prominent approach, \textbf{L2 Regularization} (or weight decay), adds the squared L2 norm of all weights to the loss: $\loss_\text{L2reg}(\params) = \loss(\params) + \lambda \sum_{l=1}^n \sqnorm*{\weightmatrix_l - \mathbf{0}}$, where $\loss(\params)$ is the original objective, $\weightmatrix_l$ the weights in layer $l$, and $\lambda$ determines regularization strength. While well-tuned L2 regularization can keep parameter magnitudes small, it often interferes with training in RL, particularly for value-based agents~\citep{DisentanglingCausesPlasticity}. \textbf{L2Init}~\citep{L2Init} modifies L2 regularization by replacing the origin with the initial weights: $\loss_\text{L2Init}(\params) = \loss(\params) + \lambda \sum_{l=1}^n \sqnorm*{\weightmatrix_l - \weightmatrix_{l,0}}$. The intuition is that initial weights are capable of quickly fitting targets, aiding plasticity preservation. Unlike targeted reset methods (Section~\ref{subsec:mitigation:targeted_resets}), L2Init does not explicitly calculate neuron utility scores. Instead, it implicitly determines the regularization strength based on the gradient magnitude: weights with large loss gradients receive less regularization, while unimportant weights are pulled toward their initial values. \textbf{2-Wasserstein regularization}~\citep{CurvatureExplainsPlasticity} addresses a limitation of L2Init, namely that initial values contain no knowledge about the learning problem. Instead of regularizing individual weights toward their initial values, Wasserstein regularization enforces similarity between the \emph{distributions} of initial and current weights using the squared 2-Wasserstein distance\footnote{The 2-Wasserstein distance is $\mathcal{W}^2_2 (p_t \mathrel{\parallel} p_0) = \sum_{l=1}^n \sum_{i=1}^d \left( \bar{w}^i_{l, t} - \bar{w}^i_{l, 0} \right)^2$ with $\bar{w}^i_{l, t}$ denoting the $i$-th largest parameter of layer $l$ at step $t$.}, yielding $\loss_{\text{2-Wass}}(\params_t, \params_0) = \loss(\params_t) + \mathcal{W}^2_2 (p_t \mathrel{\parallel} p_0)$. This is equivalent to L2Init with \emph{parameters sorted by magnitude}, a subtle change that allows larger deviations of individual weights while maintaining distributional similarity. 

Rather than constraining individual parameter values, a second family of methods controls the spectral properties of weight matrices. \textbf{Spectral Normalization (SpectralNorm)}~\citep{SpectralNorm} divides each weight matrix by its largest singular value: $\weightmatrix_l \leftarrow \weightmatrix_l / \sigma_{\max}(\weightmatrix_l)$. This makes an individual fully connected layer $K$-Lipschitz. In deep RL, SpectralNorm offers two key benefits: it curbs exploding gradients during network scaling~\citep{DeeperDeepRLSpectralNorm, OverestimationOverfittingPlasticity} and implicitly schedules the Adam learning rate~\citep{SpectralNormRLOptimization, LyleNormalizationEffectiveLR}. It also outperforms methods like clipped double Q-learning in mitigating overestimation bias and reduces dormant neurons more effectively than some targeted approaches, such as ReDO~\citep{OverestimationOverfittingPlasticity, ReDoDormantNeurons}. \textbf{Spectral Regularization}~\citep{lewandowski2024spectralregularization} takes a complementary approach by penalizing large spectral norms rather than using hard constraints on each layer. The authors observe that the rank of the parameter matrix correlates with gradient diversity (low rank implies low diversity and lost trainability). They prevent rank reduction by adding a penalty term $R_\text{spectral}(\theta_t) = \sum_{l=1}^n \left[(\sigma_{\max}(\weightmatrix_{l,t})^k - 1)^2 + \|\mathbf{b}_{l,t}\|^k\right]$, where $k$ governs how strongly large spectral norms are penalized. Both methods address spectral properties but differ in mechanism: SpectralNorm enforces a hard constraint through normalization, while Spectral Regularization applies a soft penalty during optimization.

Finally, several methods impose explicit geometric constraints on parameter norms through alternative mechanisms. \textbf{Weight Clipping}~\citep{WeightClipDeepRL} enforces that weights remain in a predefined range $[-b, b]$ after each gradient update. Here $b = \kappa \, s_l$, where $\kappa$ is a scaling hyperparameter and $s_l$ are the bounds of the uniform initialization distribution for layer $l$. The primary advantage of weight clipping over weight decay or L2Init is that it does not bias weights toward a specific point in parameter space. Instead, the parameters can move freely within bounds. However, it only works with uniform initializations, such as Kaiming Uniform. \textbf{Parseval Regularization}~\citep{parsevalNetworksOriginal} preserves row-wise orthogonality of orthogonal initialization, which offers three benefits: it implicitly regularizes all singular values to one (stronger than SpectralNorm's constraint on the largest singular value), it prevents parameter norm growth, and orthogonal weight matrices are dynamical isometries that prevent exploding or vanishing gradients~\citep{parsevalRegularization}. However, the orthogonality constraint may substantially reduce network expressivity by limiting its Lipschitz constant; this can be mitigated by learned input scaling or additional scaling layers~\citep{parsevalRegularization}.

\subsection{Feature Regularization}\label{subsec:mitigation:feature_regularization}

It is well-established that neural network layers work better with inputs that are either within specific ranges or from a standard Normal distribution. We now consider normalization in the context of plasticity loss. 

\textbf{Layer Normalization} (LayerNorm)~\citep{LayerNorm} overcomes BatchNorm's~\citep{BatchNorm} batch-size dependence and supports recurrent architectures. It normalizes a layer’s pre-activations $\activationsum_l$ using per-layer statistics before applying the nonlinearity~\citep{DisentanglingCausesPlasticity, LyleNormalizationEffectiveLR, BRO}. The transformation
$\activationsum_l \leftarrow \frac{\activationsum_l - \mathbb{E}[\activationsum_l]}{\sqrt{\operatorname{Var}[\activationsum_l] + \epsilon}} \cdot \bm{\gamma} + \bm{\beta}$
involves learnable scale $\bm{\gamma}$ and bias $\bm{\beta}$, with $\epsilon > 0$ for numerical stability.

Introduced by \citet{UnderstandingPlasticity} as a remedy for plasticity loss, LayerNorm has since shown consistent benefits~\citep{OverestimationOverfittingPlasticity, DisentanglingCausesPlasticity, LyleNormalizationEffectiveLR, BRO}. It reduces gradient covariance~\citep{UnderstandingPlasticity} and stabilizes pre-activations, mitigates overestimation bias~\citep{OverestimationOverfittingPlasticity}, reduces dormant neurons~\citep{ReDoDormantNeurons}, and curbs large gradients~\citep{OverestimationOverfittingPlasticity}. Moreover, LayerNorm enhances stability in value-based RL, akin to bounded activations~\citep{CrossQ}, by preventing unit linearization through a stabilized pre-activation distribution~\citep{DisentanglingCausesPlasticity}. It has also been shown to revive dead ReLU units by injecting non-zero gradients through normalization~\citep{LyleNormalizationEffectiveLR}, an effect confirmed in transformers~\citep{UnderstandingAndImprovingLayerNorm}.

In fact, studies with transformer networks have shown that the gradients from normalization statistics are the driving force behind LayerNorm's benefits~\citep{UnderstandingAndImprovingLayerNorm} and not the zero-mean, unit-variance activations that were  hypothesized~\citep{LayerNorm, LyleNormalizationEffectiveLR}. Moreover, \citet{LyleNormalizationEffectiveLR} demonstrated that LayerNorm introduces an implicit, parameter-norm-dependent learning rate schedule, which may be necessary for deep RL agents to learn certain behaviors.

\citet{HighVarianceRL} introduce \textbf{p-norm} to normalize the penultimate layer’s features in actor networks to prevent saturation of $\tanh$ activations in continuous control tasks. p-norm rescales the penultimate layer's features $\representation(\vecstate)$ via $\representation_{\text{norm}}(\vecstate) = \frac{\representation(\vecstate) }{\norm{\representation(\vecstate)}}$, followed by a linear layer and $\tanh$ activation. Though designed for DrQ-v2’s actor~\citep{DrQv2}, \citet{HighVarianceRL} also apply p-norm to the critic, observing notable gains in stability and performance when both networks are regularized.

Lastly, \textbf{OFN} mitigates Q-value overestimation by projecting the critic's encoder features to the unit ball~\citep{OFNDissectingRLHighUTD}, thereby decoupling the scale of Q-values from early-layer parameter norms. This prevents gradient divergence and controls parameter norm growth. The authors demonstrate that even under strong primacy bias~\citep{PrimacyBias} (overfitting to a few early samples), OFN-regularized agents can match the performance of unprimed ones. OFN also complements parameter resets, helping counteract pessimism from clipped double Q-learning~\citep{TD3OriginalPaper} to promote better exploration~\citep{OFNDissectingRLHighUTD}.

\subsection{Feature Rank Regularization}\label{subsec:mitigation:feature_rank_regularization}

Feature rank collapse frequently co-occurs with plasticity loss~\citep{UnderstandingPlasticity, L2Init, ContinualBackprop}, though the causal relationship remains unclear. Motivated by observations that under-parameterized and low-rank representations correlate with degraded performance in value-based deep RL, several algorithms explicitly target rank preservation to maintain learning capacity.

\citet{FeatureRankKumar} introduce \textbf{Direct Singular Value Regularization}, which penalizes dominant singular values of the representation matrix to encourage a more spread-out distribution, thereby increasing rank. However, this approach is highly sensitive to hyperparameter tuning and prone to producing collapsed representations. \textbf{InFeR}~\citep{InFeRUnderstandingCapacityLoss} takes a different approach, using auxiliary regression tasks with fixed random targets to indirectly encourage diverse feature learning and a higher effective rank. InFeR has shown promise in preventing plasticity loss~\citep{InFeRUnderstandingCapacityLoss} and improves representation learning in sparse-reward environments~\citep{LyleAuxiliaryTasksRepresentation}. In contrast, \textbf{DR3}~\citep{DR3FeatureRegularizer} regularizes the representation directly by penalizing large dot products between representations of consecutive states, effectively preventing feature co-adaptation. Although not designed to explicitly preserve rank, DR3 empirically prevents collapse and improves performance. Building on insights about feature similarity, \citet{AdaptiveRegularizationofRank} adaptively regularize rank based on an upper bound of the cosine similarity between consecutive state representations. Unlike the previously mentioned algorithms, their method \textbf{BEER} aims to adjust the rank according to task complexity, leading to potentially better performance and more accurate value estimates compared to methods that simply maximize rank. Unlike the methods above, which target off-policy settings, \textbf{PFO}~\citep{NoRepresentationNoTrust} addresses plasticity loss in the on-policy setting by analyzing the connection between policy updates, representation rank, and trust-region collapse in PPO~\citep{PPOOriginal}. PFO extends PPO's trust-region concept to feature space by applying an L2 penalty to prevent the pre-activation features of the current policy from drifting too far from those of the data-collecting policy.

The precise relationship between effective rank, feature rank collapse, and plasticity loss remains to be investigated. While a drop in effective rank appears to be a common pathology of agents that struggle to learn~\citep{FeatureRankKumar, L2Init, gulcehreEmpiricalStudyImplicit}, it is still debated whether maximizing rank is always beneficial, with some methods like BEER suggesting an adaptive approach is better~\citep{AdaptiveRegularizationofRank}. \citet{gulcehreEmpiricalStudyImplicit}'s study in offline RL indicates that network architecture and training hyperparameters, such as activation functions and learning rates, play a significant role in rank collapse. Furthermore, they observe a strong correlation between the number of dead units and effective rank, although without establishing a strong causal direction~\citep{gulcehreEmpiricalStudyImplicit}. In summary, while feature rank collapse is strongly associated with the loss of learning ability, its exact role and whether it is a driver or a symptom of plasticity loss are still subjects of ongoing research. 

\subsection{Activation Functions}\label{subsec:mitigation:activations}

Certain activation functions are more prone to causing dead or dormant neurons than others~\citep{ReLUDeadNeurons, ReDoDormantNeurons}, making them a natural avenue for improving network plasticity. In this section, we review various activation functions proposed in the literature to address plasticity loss, including two specifically designed to preserve it (Deep Fourier Features and Adaptive Rational Activations). For clarity, we define all functions element-wise using a scalar input $x \in \R$.

While originally proposed for image classification~\citep{CReLUoriginal}, \citet{CReLU} apply CReLU to deep RL, where  $\operatorname{CReLU} (x) = \Big[\relu (x), \relu (-x) \Big]~$ concatenates the ReLU output with its negation. They report three main benefits: improved adaptability late in training and after task switches, better gradient flow with reduced collapse, and fewer dead neurons. This last property holds since  $\operatorname{CReLU} (x) = 0$ if and only if $x=0$. In contrast, ReLU outputs zero for all $x \leq 0$.

\textbf{Parameterized Exponential Linear Units (PELU)} have been shown to outperform both ReLU and CReLU on Atari, particularly in environments with stark input distribution shifts~\citep{AdaptiveRationalActivations}. PELU generalizes the ELU activation~\citep{ELUactivation} by introducing learnable parameters~\citep{PELUactivation}, allowing the activation slope to adapt dynamically. Using learnable parameters $\alpha$ and $\beta$, it is defined as $\operatorname{PELU} (x)=\frac{\alpha}{\beta} h$ for $x \geq 0 $ and $\operatorname{PELU}(x)=\alpha\left(e^{\frac{h}{\beta}}-1\right)$ for $x <0~.$

\citet{hadamardRepresentations} introduce \textbf{Hadamard Representations} by computing the Hadamard product of two parallel hidden layers with $\tanh$ activations before the Q-function. While naive $\tanh$ activations underperform ReLU in Atari agents, Hadamard representations outperform ReLU while preserving effective rank~\citep{FeatureRankKumar} and preventing dormant neurons~\citep{ReDoDormantNeurons,hadamardRepresentations}.

\citet{DeepFourierFeatures} show that linear function approximators avoid loss of plasticity, extending this finding theoretically to deep diagonal linear networks and empirically to general deep linear networks. To balance the plasticity of linear models with the expressivity of nonlinear ones, they propose \textbf{Deep Fourier Features} using $\operatorname{Fourier} (x) = \Big[\sin (x), \cos (x) \Big]~$ as the activation function in every layer. This concatenation ensures half the units per layer are approximately linear, allowing networks with deep Fourier features to approximately embed a deep linear network with bounded error (Corollary 1~\citep{DeepFourierFeatures}).

\textbf{Adaptive Rational Activations} are defined as $\operatorname{R}(x)=\frac{\mathrm{P}(x)}{\mathrm{Q}(x)} = \frac{\sum_{j=0}^m a_j x^j}{1+\sum_{k=1}^n b_k x^k}$, where $ \mathrm{P}(x)$ and $\mathrm{Q}(x)$ are two polynomials with $m+1$ and $n$ learnable parameters. They offer two key advantages: adapting to shifts in input distribution and approximating residual connections. In practice, the denominator uses an absolute sum and $m=n+1$~\citep{AdaptiveRationalActivations}\footnote{\citet{AdaptiveRationalActivations} use $m=5$ and $n=4$ throughout their paper.}.

In summary, it appears that activation functions can help mitigate plasticity loss by reducing the number of dead units and enhancing gradient flow. However, multiple experiments have shown that plasticity loss still occurs with different activation functions~\citep{ContinualBackprop, ReDoDormantNeurons}, indicating that while specific choices may alleviate symptoms, the activation function itself is not the root cause.

\subsection{Categorical Objective Function}\label{subsec:mitigation:loss_reformulation}

\begin{figure}[h]
    \centering
    \includegraphics[width=\textwidth]{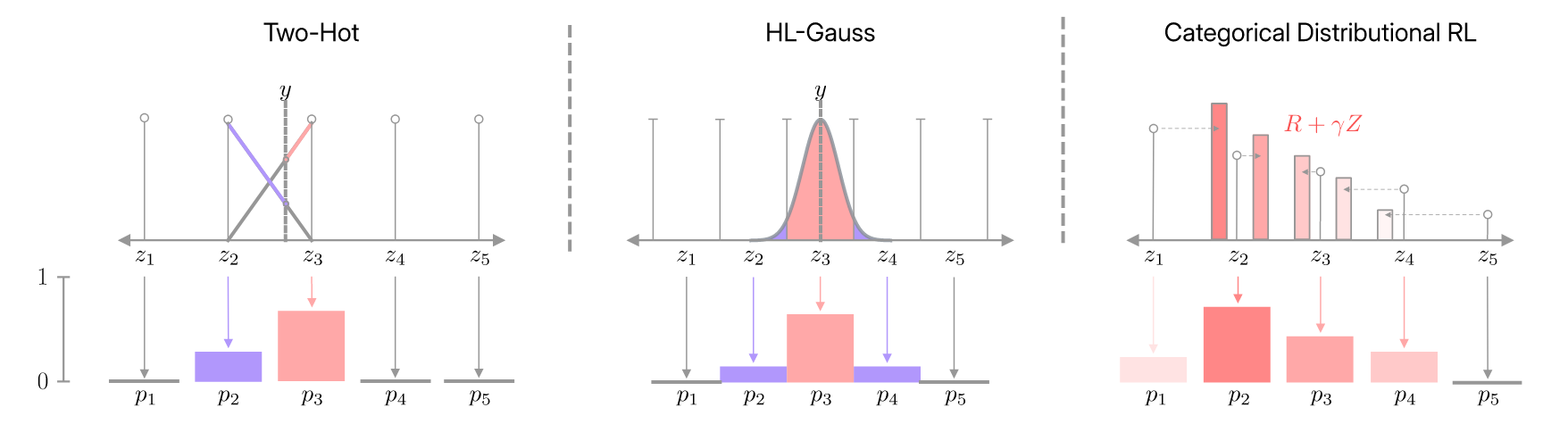}
    \Description{Overview of categorical losses.}
    \caption{\textbf{Visualization of categorical losses for deep RL}. The two-hot representation~\citep{TwoHotMuZero} proportionally assigns probability mass to the two neighboring bins of a scalar target $y$. HL-Gauss~\citep{HL-GaussOriginal} constructs a Gaussian with fixed standard deviation and integrates over each bin to obtain the corresponding probability mass. Distributional RL algorithms such as C51~\citep{C51DistributionalRL} model the full return distribution. Detailed descriptions of these methods are in Section~\ref{subsec:mitigation:loss_reformulation}. Figure taken from \citet{StopRegressing}.
    }
    \label{fig:categorical_loss_overview}
\end{figure}

It is common knowledge among deep learning practitioners that scaling network sizes is easier for classification compared to arbitrary regression tasks\footnote{Image reconstruction using per-pixel regression sidesteps common regression issues such as large errors and unstable gradients by exploiting the fact that pixel values are bounded. This enables the use of normalized objectives, which do not work for unbounded regression targets common in deep RL.}~\citep{HL-GaussOriginal, StopRegressing}. Even when regression is the actual task, reformulating the learning problem using a cross-entropy loss is often beneficial~\citep{HL-GaussOriginal, StopRegressing}. As explained in Sections~\ref{subsec:causes:objective_function} and \ref{subsec:causes:parameter_norm_growth}, regression gradients are proportional to the loss, which in turn may lead to parameter norm growth and other pathologies. One remedy is to reformulate value estimation as classification: bin a bounded reward range (usually clipped between $(-10, 10)$~\citep{C51DistributionalRL}) into discrete categories and apply a cross-entropy loss. The three most widely used categorical loss methods are C-51~\citep{C51DistributionalRL}, two-hot representations~\citep{TwoHotMuZero}, and HL-Gauss~\citep{StopRegressing}. Their differences lie in how they project a scalar target value onto the categorical bins, visualized in Figure~\ref{fig:categorical_loss_overview}.

\textbf{C-51}~\citep{C51DistributionalRL} approximates the full return distribution by directly projecting it onto $51$ categorical bins, emphasizing expressivity and distributional modeling. \textbf{Two-hot representations} offer a simpler approach, assigning probability mass only to the two nearest bins, which softens targets and reduces plasticity loss~\citep{TwoHotMuZero, DisentanglingCausesPlasticity}. \textbf{HL-Gauss} generalizes this further by first smoothing the target distribution with Gaussian noise before spreading the probability mass across multiple bins ~\citep{HL-GaussOriginal}. This results in a richer representation that better captures the ordinal structure of regression problems~\citep{StopRegressing}. All three methods use a cross-entropy loss and mitigate plasticity loss. However, they provide different trade-offs on the spectrum of expressiveness, simplicity, and training robustness.

C-51's performance gains are often attributed to enabling uncertainty quantification. However, \cite{StopRegressing} claim that the categorical loss, not distributional modeling, is the main driver of improved sample efficiency. Two-hot representations offer a lightweight alternative to distributional algorithms like C-51, but their simplicity can come at the cost of training stability~\citep{DisentanglingCausesPlasticity}. HL-Gauss enables more expressive target distributions and improved performance across tasks. Its advantages likely stem from two factors. First, distributing probability mass acts like label smoothing, which helps reduce overfitting. Second, HL-Gauss leverages the ordinal nature of regression targets, allowing for better generalization across value ranges~\citep{StopRegressing}. However, it remains unclear which categorical projection works best for a particular deep RL algorithm. For instance, \citet{StopRegressing} show that HL-Gauss performs well with discrete-control algorithms such as DQN and CQL. In contrast, \citet{BRO} find that Implicit Quantile Networks~\citep{DistributionalRLBook} outperform HL-Gauss with continuous-control SAC agents.

\subsection{Network Architectures}\label{subsec:mitigation:architectures}

Network architectures in deep RL are typically much smaller than those used in supervised learning, with agents often relying on compact MLPs~\citep{SACHaarnojaOriginal} or DQN-style CNNs~\citep{DQNOriginal, ReplayRatioBarrier}. While larger and wider networks can improve robustness to plasticity loss, optimization challenges limit their size~\citep{BiggerBetterFaster, BRO, StopRegressing}. Recent efforts to scale up deep RL models have focused on adopting residual architectures and regularization techniques, such as LayerNorm, SpectralNorm, or L2 regularization~\citep{BiggerBetterFaster, BRO, simBa}.

While the primary goal of bespoke network architectures for deep RL is often parameter scaling~\citep{prunedNetworksAreGoodNetworks,simBa}, these modifications also frequently contribute to mitigating plasticity loss. In actor-critic methods, improved architectures are mainly applied to the critic, where plasticity loss is concentrated~\citep{RevisitingPlasticityAugmentations, OverestimationOverfittingPlasticity}. Consequently, network architectures that enable stable training with a higher number of parameters often inherently improve plasticity. Many of the components utilized by the methods below were originally designed for supervised learning. With the right modifications, they have been successfully adapted to deep RL.

Several architectural approaches focus on structured sparsity and efficient parameter utilization through static architectures that do not adapt during training. Mixture of Experts (MoEs)~\citep{obandoceron2024MoE} introduce a gating mechanism that routes inputs to specialized \quoteeng{expert} networks, allowing for parameter scaling without the proportional increase in computational cost incurred by wider feedforward layers. BroNet~\citep{BRO} and SimBa~\citep{simBa} utilize stacked residual blocks~\citep{ResNet} in the critic of SAC agents to efficiently scale parameters while maintaining stable training. Both are inspired by transformer architectures and share common building blocks, such as LayerNorm and residual connections. Compared to BroNet, SimBa utilizes an input normalization layer called RSNorm, which helps in avoiding overfitting when scaling up the network~\citep{simBa}. Both BroNet and SimBa can be used as replacements for standard critic architectures in agents like SAC. Unlike BRO, SimBa does not necessitate further algorithmic modifications like using a distributional loss and weight decay~\citep{BRO,simBa}. MoEs, SimBa, and BroNet all enable parameter scaling and help mitigate plasticity loss. MoEs achieves this by selectively activating parts of the network in the penultimate layer, which contains most of the parameters of a deep RL agent. BroNet and SimBa instead rely on residual connections and normalization techniques to enable stable parameter scaling without overfitting and training instabilities.

Other methods allow for dynamic manipulation of the network's structure during training. Network Pruning~\citep{prunedNetworksAreGoodNetworks} is based on the observation that networks often under-utilize parameters. 
It applies gradual magnitude pruning to remove the least important connections, enhancing performance and enabling network scaling while improving gradient covariance and other plasticity metrics (See Section~\ref{subsec:causes:first_order_effects}). Neuroplastic Expansion (NE)~\citep{neuroplasticExpansion} dynamically adjusts network topology both by adding new connections based on gradient norms and pruning dormant neurons. The topology adjustment phase is interleaved with a consolidation phase to prevent catastrophic forgetting. In contrast to pruning, NE directly maintains plasticity by adding new connections while addressing catastrophic forgetting through the consolidation phase~\citep{neuroplasticExpansion}. Both pruning and NE aim to make network usage more efficient, but pruning primarily reduces redundant parameters, while NE actively adjusts the network to adapt to new information.

\subsection{Distillation}\label{subsec:mitigation:distillation}

Distillation algorithms aim to transfer knowledge from a \emph{teacher network} to a \emph{student network}, mitigating negative side effects during training~\citep{NNDistillation}. \textbf{ITER}~\citep{ITERNonstationarity} periodically distills the actor and critic networks of PPO agents to address plasticity loss. By updating the student in parallel with the teacher's RL training, ITER avoids requiring a replay buffer in on-policy algorithms like PPO. \textbf{Hare \& Tortoise Networks}~\citep{HareTortoiseNetworks} combine distillation with periodic resets using a dual architecture: a \quoteeng{\emph{hare}} network rapidly learns via SGD while a \quoteeng{\emph{tortoise}} network slowly integrates information through an exponential moving average of the hare's parameters. The hare is periodically reset to the tortoise's parameters, enabling frequent resets without losing accumulated knowledge and providing a mechanism to escape suboptimal local minima for the hare.

\subsection{Optimization}\label{subsec:mitigation:optimization}

Optimization in deep RL differs from supervised learning due to non-stationary data and shifting targets that violate i.i.d. assumptions. Value-based RL algorithms perform fixed-point iteration, approximated by stochastic gradient descent~\citep{SuttonBartoRLIntroduction}, which complicates optimization. While momentum-based optimizers such as Adam~\citep{AdamOptimizer} work reasonably well by automatically selecting step sizes, they often require adjustments~\citep{adamRel, ResettingAdamMoments} like a much larger stability parameter $\epsilon$~\citep{cleanRL, ReDoDormantNeurons} to prevent moment estimate divergence. This has motivated RL-specific optimizers, such as Adam-Rel \citep{adamRel} and OPEN \citep{OPENLearnedOptimization}, which are presented in this section.

OPEN~\citep{OPENLearnedOptimization} meta-learns an optimizer to address deep RL’s challenges: non-stationarity, plasticity loss, and exploration. The gradient update uses a small RNN with learned stochasticity, meta-trained to optimize final return. By conditioning on features like neuron dormancy and network depth, it becomes naturally robust to RL's non-stationarity. OPEN uses parameter-space noise to reawaken dormant neurons~\citep{ReDoDormantNeurons} similar to \citet{AddressingPlasticityCatastrophicForgetting}, effectively aiding exploration and preventing plasticity loss. This explicit focus on RL difficulties yields improved performance across environments compared to other meta-learning approaches. On the other hand, Sharpness-Aware Minimization (SAM)~\citep{SharpnessAwareMinimization} improves generalization by seeking flat minima, which may enhance plasticity by reducing loss landscape curvature~\citep{PLASTIC, UnderstandingPlasticity}. Unlike other optimizers, SAM explicitly perturbs gradients to identify low-curvature regions, requiring two gradient computations per iteration~\citep{SharpnessAwareMinimization}. While OPEN uses parameter noise to maintain plasticity, SAM perturbs gradients to find flat minima~\citep{SharpnessAwareMinimization}.

A separate line of work focuses on modifying optimizers for deep RL. Adam-Rel~\citep{adamRel} resets Adam's internal step count every time a non-stationarity occurs, enabling rapid adaptation of moment terms to shifts in gradient statistics~\citep{adamRel}. Moment resetting~\citep{ResettingAdamMoments} is a related technique that reinitializes momentum buffers at each target network update to prevent outdated moments from contaminating updates. In off-policy RL, target network updates can increase gradient norms disproportionately between first and second moments \footnote{PyTorch uses default values of $\beta_1 = 0.9$ for the first moment and $\beta_2 = 0.999$ for the second moment
~\citep{PytorchCitation}.}, potentially causing training divergence~\citep{UnderstandingPlasticity}. While both methods address the mismatch between stationarity assumptions and non-stationary RL dynamics, moment resetting is more severe. In on-policy RL, it may discard useful optimization information that Adam-Rel preserves~\citep{adamRel}.

\subsection{Other Methods}\label{subsec:mitigation:other_regularizers}

In this section, we discuss methods relevant to plasticity loss that did not fit into previous categories. Many are well-known regularization techniques, such as specific types of representations, input/target scaling, or hyperparameter selection. While they were not initially designed to mitigate plasticity loss, they nonetheless affect it. For instance, \textbf{input and target scaling} can reduce input or target non-stationarity. Both techniques have been around for a while~\citep{PPOOriginal} and may yield environment-dependent performance benefits~\citep{ImplementationMattersForPolicyGradients, WhatMattersForOnPolicy, PPOImplementationDetails}. On the hyperparameter front, AdaQN~\citep{AdaQN} automatically selects hyperparameters online by training an ensemble of Q-networks with different configurations and selecting a shared target network based on recent loss values, enabling adaptation as optimal hyperparameters shift across training stages~\citep{hyperparameterconsistency}.

\textbf{Data Augmentations} have driven impressive sample efficiency gains in pixel-based control. Their success is conventionally attributed to improved representation learning~\citep{RADRLAlgorithm, DrQDrQeps, DrQv2}. However, \citet{RevisitingPlasticityAugmentations} provide evidence that data augmentations actually mitigate plasticity loss during critic training by preventing an early drop in active units. Building on this insight, they propose increasing the gradient steps per environment step once the early stage of plasticity loss has passed. Borrowing from vector-quantized generative models~\citep{VQVAE}, \citet{DiscreteRepresentations} find that \textbf{Sparse Representations} generalize better and adapt faster to environmental non-stationarities when training world models. They attribute these results to the sparse, binary nature of one-hot embeddings: one-hot representations substantially outperform quantized ones despite identical information content~\citep{DiscreteRepresentations}.

\subsection{Combined Methods}\label{subsec:mitigation:combined_methods}

\begin{figure}[t]
\centering
\includegraphics[width=0.8\textwidth]{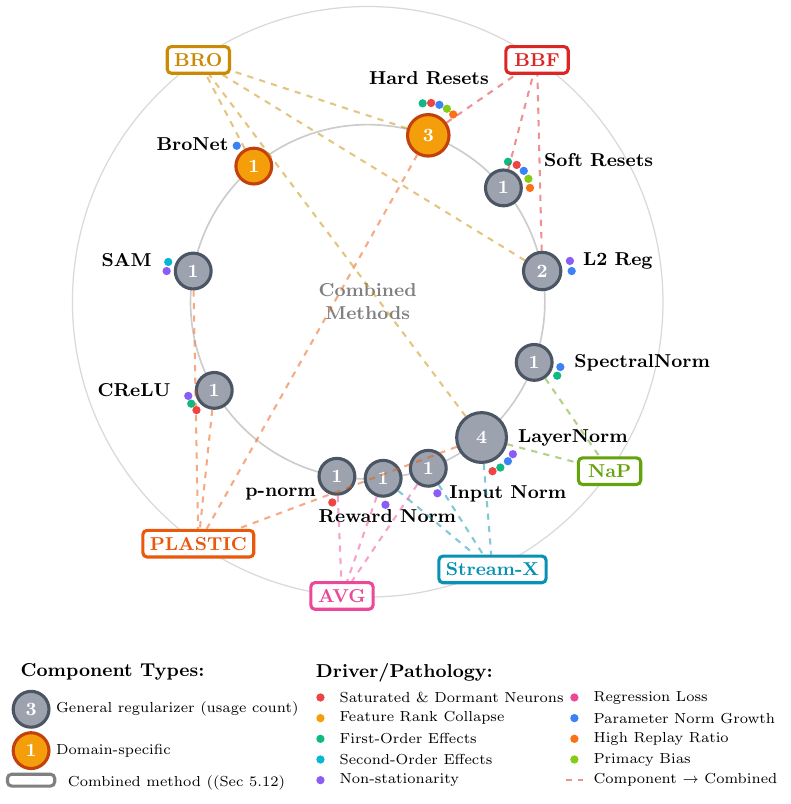}
\Description{Overview of combined plasticity methods and their components.}
\caption{Combined plasticity loss mitigation methods from Section 5.12. \textbf{Inner ring}: Component methods, colored by type (blue = general regularizer, orange = domain-specific), sized by usage frequency. Number inside node indicates how many combinations use this component. \textbf{Outer ring}: Six published combined methods. Colored dots indicate causes addressed (see legend). Dashed lines connect components to combinations. LayerNorm is most frequently used (4/6), followed by Hard Resets (3/6) and L2 Regularization (2/6), demonstrating that effective combinations leverage general regularizers across categories.}
\label{fig:plasticity-combined-methods}
\end{figure}

Since the exact causes of plasticity loss remain unclear and likely involve multiple interacting factors~\citep{UnderstandingPlasticity, DisentanglingCausesPlasticity}, integrating different mitigation strategies that target specific symptoms is a sensible approach. Consequently, many successful algorithms employ multiple complementary regularizers. The first popular combination is LayerNorm + L2 regularization, introduced by \citet{UnderstandingPlasticity}. Subsequently, \citet{OverestimationOverfittingPlasticity} examine well-performing combinations for SAC agents and found that Resets + L2 regularization, as well as LayerNorm + Resets, work particularly well. In the following, we will summarize particularly well-performing combinations that have achieved strong results.

Effective combinations often explicitly address multiple potential drivers, such as input and target non-stationarity, while controlling pathologies like growing parameter norms simultaneously. One such algorithm, \textbf{PLASTIC}~\citep{PLASTIC}, combines the CReLU activation function~\citep{CReLU} with sharpness-aware optimization (SAM)~\citep{SharpnessAwareMinimization}, LayerNorm~\citep{LayerNorm}, and resets~\citep{PrimacyBias} to achieve strong performance on both Atari-100k and the DeepMind Control Suite. Another combination, Normalize-and-Project (\textbf{NaP})~\citep{LyleNormalizationEffectiveLR}, pairs LayerNorm with a projection step similar to SpectralNorm, where weights are projected onto a ball with radius $\rho$ instead of the unit ball:  $\weightmatrix_l \leftarrow \rho \weightmatrix_l / \norm{\weightmatrix_l}$. This combination removes LayerNorm's implicit parameter-norm-dependent learning rate schedule, but potentially necessitates the introduction of an explicit schedule instead. \citet{LyleNormalizationEffectiveLR} find that simple linear decay proportional to parameter norm growth suffices. 

Achieving strong performance often requires scaling up network capacity while deploying multiple plasticity-preserving techniques simultaneously. \textbf{BBF}~\citep{BiggerBetterFaster} achieves state-of-the-art mode-free performance on Atari-100k by employing a deeper ResNet-based architecture (IMPALA-CNN~\citep{IMPALA}). Furthermore, BBF uses hard resets for the network head with soft resets for the encoder, and employs weight decay to enable stable training at high replay ratios. \textbf{BRO}~\citep{BRO} set the then state-of-the-art for proprioceptive continuous control by combining full network resets with LayerNorm and weight decay. Additionally, BRO incorporates a bespoke residual network architecture, optimistic exploration, and a quantile network for the critic, further enhancing sample efficiency.

Lastly, both \textbf{AVG}~\citep{AVGPolicyGradientWithoutBatch} and \textbf{Stream-X}~\citep{streamingRLFInallyWorks} tackle streaming RL in settings where agents process transitions one by one from a single environment, precluding batch updates. Due to the high variance in this regime, both methods employ observation and reward/error normalization. Notably, scaling reward values may reduce plasticity loss by avoiding regression to large targets (cf. Section~\ref{subsec:causes:objective_function}), which is associated with harmful parameter norm growth (cf. Section~\ref{subsec:causes:parameter_norm_growth}). AVG uses p-norm~\citep{HighVarianceRL} for additional regularization, while Stream-X employs a custom optimizer, sparse initialization, and LayerNorm to maintain plasticity throughout training. Both methods achieve performance competitive with traditional deep RL algorithms on high-dimensional continuous and discrete control benchmarks.

%% file: sections/6discussion_directions.tex
\section{Current State and Future Directions}\label{sec:discussion_directions}

The field of plasticity loss has made substantial empirical progress in recent years, though significant gaps in understanding remain. In this section, we synthesize the current state of knowledge and identify the strength of evidence for particular aspects.

We know that multiple interventions robustly improve performance~\citep{UnderstandingPlasticity, OverestimationOverfittingPlasticity, SpectralNormRLOptimization, StopRegressing}: resets~\citep{ShrinkPerturb, PrimacyBias}, LayerNorm~\citep{LayerNorm}, SpectralNorm~\citep{SpectralNorm}, and categorical losses~\citep{HL-GaussOriginal}. In deep RL, networks suffering from plasticity loss show multiple measurable pathologies that correlate with reduced performance: dormant neurons~\citep{ReDoDormantNeurons}, rank collapse~\citep{FeatureRankKumar, NoRepresentationNoTrust}, growing parameter norms~\citep{DisentanglingCausesPlasticity, lewandowski2024spectralregularization}, and gradient pathologies~\citep{OFNDissectingRLHighUTD, ACEOffPolicyActorCriticCausallyAware, lewandowski2024spectralregularization}. Both input and target non-stationarity occur ubiquitously in this domain~\citep{ReDoDormantNeurons, PLASTIC, HareTortoiseNetworks}, though quantifying its severity remains challenging.

Despite the empirical progress, we lack a fundamental understanding of why these interventions work. Resets restore learning ability~\citep{PrimacyBias}, but whether they work by eliminating dormant neurons, reducing parameter norms, improving optimization dynamics, or some combination remains unclear. LayerNorm~\citep{LayerNorm} is among the most effective regularizers, yet we cannot definitively state which effect drives its benefits: maintaining normalized activations~\citep{LayerNorm, LyleNormalizationEffectiveLR}, inducing implicit learning rate schedules~\citep{LyleNormalizationEffectiveLR}, or preventing dead neurons through normalization gradients~\citep{UnderstandingAndImprovingLayerNorm}.

Currently, the field has identified correlations but rarely established causation. Figure~\ref{fig:plasticity_loss_model} presents plausible connections between potential factors contributing to plasticity loss, but these relationships represent hypotheses rather than proven causal links. For example, target non-stationarity correlates with dormant neurons~\citep{ReDoDormantNeurons}, but this finding does not constitute a rigorous causal relationship. Beyond gaps in understanding, the field faces methodological challenges that limit the generalizability and reproducibility of research on plasticity loss. In the following subsections, we examine current methodological issues that constrain progress and outline priority research directions that could advance both our understanding and practical mitigation of plasticity loss.

\subsection{Current Methodological Issues}\label{subsec:discussion_directions:current_issues}

Research on plasticity loss has traditionally focused on Atari for discrete control and the DeepMind Control Suite (DMC) for continuous control. However, broadening the evaluation reveals that environments exhibit different pathologies at different severities: while DMC's Dog environment causes gradient explosion in SAC agents, MetaWorld environments suffer from greater parameter norm growth~\citep{OverestimationOverfittingPlasticity}. Intervention effectiveness varies accordingly: LayerNorm~\citep{LayerNorm} substantially improves performance on Dog by stabilizing gradients, yet it is actively harmful when applied on MetaWorld~\citep{OverestimationOverfittingPlasticity}. Similar effects can be seen across algorithms: While ReDo helps for the off-policy algorithm DQN on Atari~\citep{ReDoDormantNeurons}, it is outperformed by other parameter regularizers for the on-policy algorithm PPO~\citep{studyPlasticityOnPolicyRL}. This variability highlights a critical gap: We cannot yet predict which environment properties will cause severe plasticity loss and which interventions will prove effective for a given task. Without this predictive capability, practitioners must resort to exhaustive trial-and-error testing.  Systematic evaluation across diverse benchmarks could identify measurable environment properties that predict the severity of plasticity loss, moving the field from reactive to proactive algorithm design. Such evidence-based regularizer selection would also facilitate testing for unintended side effects of popular methods.

This systematic evaluation, however, depends on establishing consensus on measurement. The field currently lacks agreement on how to measure plasticity loss across different settings, making it difficult to compare interventions across studies. Researchers employ various metrics, including neuron dormancy ratios, effective rank, and gradient norms, but none are used consistently or comprehensively. This inconsistency prevents systematic meta-analysis of which algorithms work best under which conditions, obscuring the ability to identify universal principles underlying plasticity loss. We believe that a standardized evaluation protocol should prioritize the metrics laid out in Table~\ref{tab:standardized_metrics}. Notably, we omit feature rank and neuron dormancy metrics. This is because we believe neuron dormancy to be a downstream effect of optimization issues that can be captured better by measuring gradient norms~\citep{ACEOffPolicyActorCriticCausallyAware}, and parameter norms to be a good proxy indicator for rank issues~\citep{NoRepresentationNoTrust, lewandowski2024spectralregularization}.

\begin{table}[h]
\centering
\begin{tabular}{lll}
\toprule
\textbf{Metric Category} & \textbf{Specific Measurements} & \textbf{Rationale} \\
\midrule
\textbf{Parameter norms} & Parameter norms per layer & Easy to measure; first-stop diagnostic \\
& & tool for optimization issues \\
\midrule
\textbf{Gradient metrics} & L1/L2 norms per layer, gradient spectrum & Foundations of gradient-based learning; \\
& & likely precede other pathologies \\
\midrule
\textbf{Curvature} & Largest Hessian eigenvalue & Fundamental insights into optimization \\
& & landscape \\
\bottomrule
\end{tabular}
\caption{Proposed standardized metrics for plasticity loss evaluation.}
\label{tab:standardized_metrics}
\end{table}

\subsection{Understanding Mechanisms (Highest Priority)}\label{subsec:discussion_directions:understanding_mechanisms}
Establishing causal mechanisms behind plasticity loss represents the field's most critical gap. Without a mechanistic understanding, we develop techniques through trial and error rather than principled design. Making progress requires controlled studies that isolate individual factors and establish causal relationships, rather than just benchmarking final performance. We believe that three questions must be answered to gain a deeper understanding:
\begin{enumerate*}[label=(\alph*)]
    \item understanding why successful regularizers work,
    \item identifying the causal relationships between observed pathologies, and
    \item developing methods to quantify non-stationarity.
\end{enumerate*}

Despite strong empirical evidence that certain regularizers consistently mitigate plasticity loss, the underlying mechanisms of these regularizers remain unclear. LayerNorm is perhaps the most successful intervention, yet we cannot definitively state whether its benefits arise from maintaining normalized activations, providing gradients through normalization statistics, inducing implicit learning rate schedules, or reviving dead neurons~\citep{UnderstandingAndImprovingLayerNorm, UnderstandingPlasticity, DisentanglingCausesPlasticity, LyleNormalizationEffectiveLR}. Similarly, L2 regularization and L2Init~\citep{L2Init} both improve plasticity, but it remains unclear whether they primarily work by controlling parameter norm magnitudes or through implicit functional regularization. SpectralNorm's benefits could stem from enforcing Lipschitz constraints, from implicit learning rate effects, or from both mechanisms operating simultaneously~\citep{DeeperDeepRLSpectralNorm, SpectralNormRLOptimization, OverestimationOverfittingPlasticity}. The field needs mechanistic studies that isolate and test individual hypotheses rather than focusing on benchmark performance.

Figure~\ref{fig:plasticity_loss_model} depicts a plausible causal chain: non-stationarity plus large-mean regression leads to gradient instability, which causes parameter norm growth, which produces pathologies including sharp landscapes, saturated units, and rank collapse, ultimately reducing performance. However, critical links remain poorly understood. The connection between parameter norms and sharpness exemplifies this gap. \citet{DisentanglingCausesPlasticity} find an empirical correlation between parameter norms and maximum Hessian eigenvalues, but do not establish a causal relationship. Whether this correlation arises fundamentally during the training of neural networks with non-stationarity or whether it is an artifact of specific architectures and activation functions remains unknown. Rank collapse and dormant neurons correlate strongly~\citep{gulcehreEmpiricalStudyImplicit} in offline RL, but the causal direction remains unclear as well. Both may be downstream consequences of gradient collapse from parameter norm growth rather than causing each other.

While non-stationarity clearly contributes to plasticity loss, the precise mechanisms underlying this phenomenon remain elusive. In deep RL, two primary forms of non-stationarity occur: target non-stationarity from bootstrapping (Section~\ref{subsec:preliminaries:rl}) and input non-stationarity from policy changes~\citep{PLASTIC}. Evidence from different studies conflicts, with some emphasizing target shifts and others highlighting changes in input distribution~\citep {ITERNonstationarity, ReDoDormantNeurons, PLASTIC}. More fundamentally, the field lacks methods to quantify the degree of non-stationarity in a given learning problem. Such quantification is necessary for principled regularization selection: stronger non-stationarity should require stronger regularization, but without measurement tools, this principle cannot be implemented. Additionally, plasticity loss occurs in both classification and regression under non-stationarity~\citep{ITERNonstationarity, HareTortoiseNetworks, DisentanglingCausesPlasticity}, yet these tasks have different loss landscapes and optimization dynamics. Identifying the common factors linking non-stationarity to plasticity loss across settings remains an open question.

\subsection{Connections to Established RL Issues}\label{subsec:discussion_directions:connections_to_other_issues}
Research on plasticity loss has revealed that many classical RL problems respond to general neural network regularization rather than domain-specific algorithmic solutions. Overestimation bias has motivated extensive work on double Q-learning and pessimistic value estimation~\citep{DoubleQLearning,TD3OriginalPaper}, yet recent benchmarks suggest plasticity interventions address it more effectively~\citep{OverestimationOverfittingPlasticity, AdaptiveRationalActivations, OFNDissectingRLHighUTD}. Exploration may be driven partially by frequent action changes from gradient-based learning instead of explicit mechanisms like $\epsilon$-greedy~\citep{PolicyChurn, DrMDormantRatio}. Most strikingly, Baird's counterexample demonstrated that certain combinations of bootstrapping and function approximation must diverge; yet, LayerNorm and L2 regularization stabilize this exact case~\citep{PQNSimplifyingTemporalDifference}. Recent advances in normalization have eliminated the need for perceived cornerstone stability techniques, such as target networks, alongside BatchNorm or LayerNorm. Despite this simplification, performance is maintained or improved~\citep{CrossQ, PQNSimplifyingTemporalDifference}.

These findings do not prove that all deep RL problems reduce to optimization issues, but they warrant a systematic revisiting of classical problems. Consider the deadly triad: bootstrapping, function approximation, and off-policy learning cannot be safely combined~\citep{DeadlyTriad}. Is this instability truly inherent, or does it stem from insufficient regularization? Similarly, passive learning exhibits degraded performance when learning from data generated by other policies~\citep {DifficultyOfPassiveLearning}. Does this reflect a fundamental off-policy difficulty or network optimization issues due to distribution shift? Testing whether plasticity methods resolve these classical failures would clarify which problems are algorithmic versus optimization-based, with direct implications for when domain-specific approaches are required versus where standard deep learning regularizers suffice.

\subsection{Theory Development}\label{subsec:discussion_directions:theory}

Despite substantial empirical progress, we lack theoretical frameworks to predict when plasticity loss will occur or which interventions will work for a given problem. Rather than aspiring to general frameworks, we identify specific theoretical advances that would directly inform practice.

Can we bound the rate at which plasticity is lost under non-stationarity as a function of measurable properties such as target shift magnitude, batch size, and learning rate? Such bounds would predict the severity of plasticity loss a priori, rather than discovering it through extensive experimentation. This connects to a more fundamental question about the relationship between optimization dynamics and plasticity: how does maintaining plasticity throughout training affect total sample complexity to reach target performance? Formalizing this relationship would quantify the practical cost of plasticity loss and justify the computational overhead of plasticity-preserving algorithms. A critical missing piece is the formal connection between parameter norm growth and loss landscape curvature established empirically by \citet{DisentanglingCausesPlasticity}. How exactly are parameter norm growth and the maximum eigenvalue of the Hessian connected? As previously mentioned, current understanding relies on empirical correlation rather than rigorous derivation. Establishing this connection theoretically would explain a central mechanism linking non-stationarity to optimization difficulty. Additionally, formalizing the parameter norm-curvature relationship may reveal whether it holds universally across architectures or depends on specific design choices, such as activation functions, architecture, or loss.

Categorical losses typically outperform regression for plasticity preservation~\citep{HL-GaussOriginal, UnderstandingPlasticity, StopRegressing}. However, this advantage is not universal; in environments like Atari's Phoenix and Alien, MSE substantially outperforms classification reformulations. We lack a theoretical justification for why and when categorical losses excel. Regression gradients scale with prediction error, potentially causing parameter norm explosion under large-mean targets, which are common in value-based RL. Cross-entropy gradients, in contrast, remain bounded even for large target values. Does this bounded gradient property alone explain the benefits of categorical losses? Or do other factors contribute substantially, such as the implicit label smoothing from distributing probability mass by methods like HL-Gauss~\citep{HL-GaussOriginal}? Formalizing these mechanisms would clarify when categorical reformulations help and guide their design for other continuous prediction problems beyond value functions.

%% file: sections/7conclusion.tex
\section{Conclusion}\label{sec:conclusion}

Plasticity loss represents a fundamental challenge in deep RL, where networks progressively lose their ability to learn despite remaining capacity. This survey consolidated fragmented definitions into a unified formulation (Definition~\ref{def:plasticity_loss}) and developed the first systematic taxonomy of factors and mitigation strategies encompassing approximately fifty methods across twelve categories (Figures~\ref{fig:plasticity-combined-methods} and~\ref{fig:plasticity-bipartite}). Our analysis reveals a central finding: general regularization techniques tend to outperform domain-specific plasticity interventions across diverse benchmarks. The most successful agents often combine these well-established techniques rather than relying on novel RL-specific mechanisms. This pattern extends beyond plasticity loss itself, as general regularizers also mitigate overestimation bias, improve exploration, and stabilize training where specialized algorithms were previously thought necessary (Sections~\ref{subsec:discussion_directions:connections_to_other_issues}).

Given these findings, we recommend that practitioners encountering plasticity loss should start with general regularization techniques such as LayerNorm and SpectralNorm. These methods offer two advantages: they frequently match or exceed the performance of specialized approaches, and they benefit from battle-tested, efficient implementations in standard deep learning frameworks. When additional intervention is needed, soft resets provide an effective next step due to their simplicity and broad applicability across warm-starting, on-policy, and off-policy settings.

However, fundamental understanding remains limited despite empirical progress. Plasticity loss manifests through multiple measurable pathologies: dormant neurons, rank collapse, parameter norm growth, and gradient issues. Non-stationarity clearly contributes as well, but neither its degree nor which environments trigger severe plasticity loss can be predicted. Compounding these gaps, the field lacks standardized evaluation protocols, with researchers designing bespoke experiments tracking different metrics. Current work also focuses on Atari and the DeepMind Control Suite, where the effectiveness of methods varies across tasks (Sections~\ref{sec:causes} and \ref{subsec:discussion_directions:current_issues}).

The highest priority for future research is establishing causal mechanisms. In Figure~\ref{fig:plasticity_loss_model}, we synthesize the existing drivers and pathologies of plasticity loss into a hypothetical causal model. Because these associations are empirical rather than theoretically grounded, the causal relationships remain unclear. For example, the correlation between parameter norm growth and loss landscape curvature is known, but their directional dependence is still unknown. Understanding why successful regularizers work, such as which of LayerNorm's multiple effects drive its benefits, remains unresolved. General regularization methods can mitigate classical RL problems, such as overestimation bias and the deadly triad. Consequently, these issues may stem from optimization pathologies when using RL with deep learning rather than fundamental algorithmic limitations. Systematically revisiting these classical problems could clarify which challenges genuinely require domain-specific solutions (Sections~\ref{subsec:discussion_directions:understanding_mechanisms}, \ref{subsec:discussion_directions:theory}). We believe that success for the field will be measured not by novel algorithms but by predictive theories: frameworks that explain when plasticity loss will occur based on measurable environment and training properties, why specific regularizers succeed, and how to select interventions based on problem characteristics rather than exhaustive search. We hope this survey provides a foundation for moving toward such principled, mechanistic understanding.